\documentclass[final,5p,times,twocolumn,authoryear]{elsarticle}

\usepackage[utf8]{inputenc}
\usepackage[T1]{fontenc}
\usepackage[english,vietnamese]{babel}
\usepackage{booktabs}
\usepackage{epsfig}

\usepackage{makecell}

\usepackage{xcolor}  % text color
\usepackage{hyperref} % show the hyper-ref
\usepackage{caption}
\usepackage{subcaption}
\usepackage{amssymb}
\usepackage{multirow}
\usepackage{pdflscape}
\usepackage{amsfonts}
\usepackage{wrapfig}
\usepackage{adjustbox}
\usepackage{algorithm}
\usepackage{amsmath,amssymb,bm}
\usepackage{array}
\usepackage{enumitem}

\hypersetup{
    colorlinks = true
}

\AtBeginDocument{%
    \hypersetup{
        colorlinks = true,
        citecolor = red,
        linkcolor = red,
        urlcolor  = black
    }%
}

\newproof{proof}{Proof}

\newcolumntype{C}[1]{>{\centering\arraybackslash}m{#1}}
\usepackage[nodisplayskipstretch]{setspace}
\usepackage{tabularx}
\usepackage{graphicx}
\usepackage{bm}
\allowdisplaybreaks

\let\cite\citep

\usepackage{balance}

\journal{Neurocomputing}
\begin{document}
\selectlanguage{english}
\setlength{\belowdisplayskip}{3pt} 
\setlength{\belowdisplayshortskip}{3pt}
\setlength{\abovedisplayskip}{3pt} 
\setlength{\abovedisplayshortskip}{3pt}

\begin{frontmatter}

\title{FSS-UBrain: Multi-region Few-Shot Brain Tumor MRI Segmentation}
% \title{FS-SwinPResU: An Efficient Hybrid Framework for Precise Brain Tumor MRI Segmentation}
% List of Authors
\author[inst1]{Truong Viet Vu}\ead{vu.2274802011045@vanlanguni.vn}
\author[inst1]{Nguyen Phuc Nguyen}\ead{nguyen.2274802010586@vanlanguni.vn}
\author[inst1]{Dang Thi Thu Hang}\ead{hang.2274801040002@vanlanguni.vn}
\author[inst2]{Tran Thien Thanh}\ead{thanh.tran@ut.edu.vn}
\author[inst1]{Vo Nguyen Quoc Bao}\ead{bao.vnq@vlu.edu.vn}
\author[inst1]{Nguyen Thai Anh}\ead{anh.nt@vlu.edu.vn}
\author[inst1]{Ngo Hoang Tu\corref{cor1}}\ead{tu.nh@vlu.edu.vn}
\cortext[cor1]{Corresponding author.}
% Define Affiliation
\affiliation[inst1]{organization={Faculty of Information Technology, Van Lang School of Technology, Van Lang University, Ho Chi Minh City 70000}, country={Vietnam}}
\affiliation[inst2]{organization={Institute of Information Technology and Electrical-Electronics Engineering, Ho Chi Minh City University of Transport, Ho Chi Minh City 70000}, country={Vietnam}}
% Corresponding author
% \affiliation[corres]{country={Corresponding author: Ngo Hoang Tu (tu.nh@vlu.edu.vn)}}

%%%%%%%%%%%%%%%%%%%%%%%%%  ABSTRACT  %%%%%%%%%%%%%%%%%%%%%%%%%%%%%%%
\begin{abstract}
Accurate delineation of whole tumor (WT), tumor core (TC), and enhancing tumor (ET) from multimodal magnetic resonance imaging remains challenging under limited annotation, cross-cohort variation, and severe target sparsity. We propose FSS-UBrain, a region-wise one-shot segmentation framework that uses a labeled positive support slice to condition binary query segmentation separately for WT, TC, and ET. Support-derived foreground and background descriptors guide query-feature adaptation, bottleneck interaction, decoder-side reconstruction, and boundary refinement. Episodic training additionally incorporates hard-negative and fully negative queries with empty-query regularization to suppress spurious foreground activation when the selected region is absent. Although inference operates on two-dimensional support--query slice pairs, checkpoint selection, threshold calibration, and final evaluation are performed after volumetric reconstruction. FSS-UBrain is evaluated on a held-out BraTS 2020 split and under target-supported cross-cohort protocols on BraTS 2023 and BraTS-Africa. Cases used as target support are excluded from the query cohorts, and no target-domain fine-tuning or test-time parameter updates are performed. On BraTS 2020, FSS-UBrain achieves volumetric Dice scores of 89.82\%, 82.14\%, and 77.42\% for WT, TC, and ET, respectively, with corresponding 95th-percentile Hausdorff distance (HD95) values of 11.12, 9.01, and 4.46~mm. It also achieves the highest mean Dice and lowest finite-pair mean HD95 point estimates on BraTS 2023 and BraTS-Africa among the compared few-shot methods. These findings support target-conditioned few-shot segmentation while highlighting sensitivity to support selection and cohort-specific variation.
\end{abstract}


\end{frontmatter}

%%%%%%%%%%%%%%%%%%%%%%%%%%%%%%%%%%%%%%%%%%%%%%%%%%%%%%%%%%%%%%%%%%%%%%%%
\section{Introduction}
\label{Sect:1}
Brain and central nervous system tumors continue to impose a substantial clinical burden despite advances in diagnosis and treatment \cite{zhao2025global,huang2023disease}. In routine neuro-oncology workflows, accurate delineation of tumor subregions, including whole tumor (WT), tumor core (TC), and enhancing tumor (ET), from multimodal magnetic resonance imaging (MRI) is critical for diagnosis, surgical planning, radiotherapy targeting, and longitudinal monitoring. However, manual annotation is labor-intensive, time-consuming, and prone to inter-observer variability, motivating the development of reliable automated segmentation systems \cite{despotovic2015mri,menze2018multimodal,bakas2018identifying}.

\begin{figure*}[t!]
\centering
% \vspace{-0.5cm}
\begin{subfigure}{.33\linewidth}
  \centering
  % include first image
  \includegraphics[width=\linewidth]{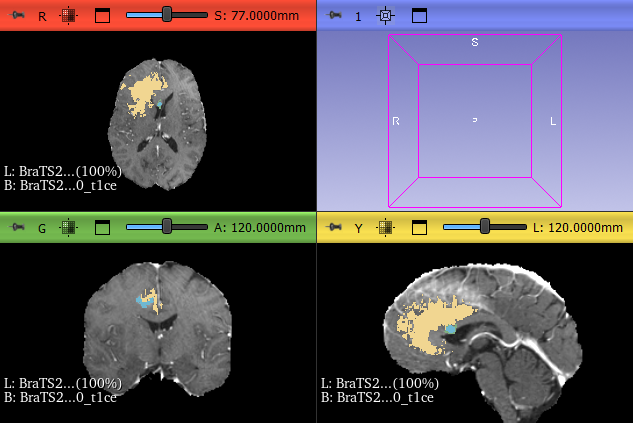}  
  \caption{BraTS20 \texttt{BraTS20\_Training\_030}}
  \label{fig:sub-11-Config1}
  % \vspace{0.2cm}
\end{subfigure}
\begin{subfigure}{.33\linewidth}
  \centering
  % include second image
  \includegraphics[width=\linewidth]{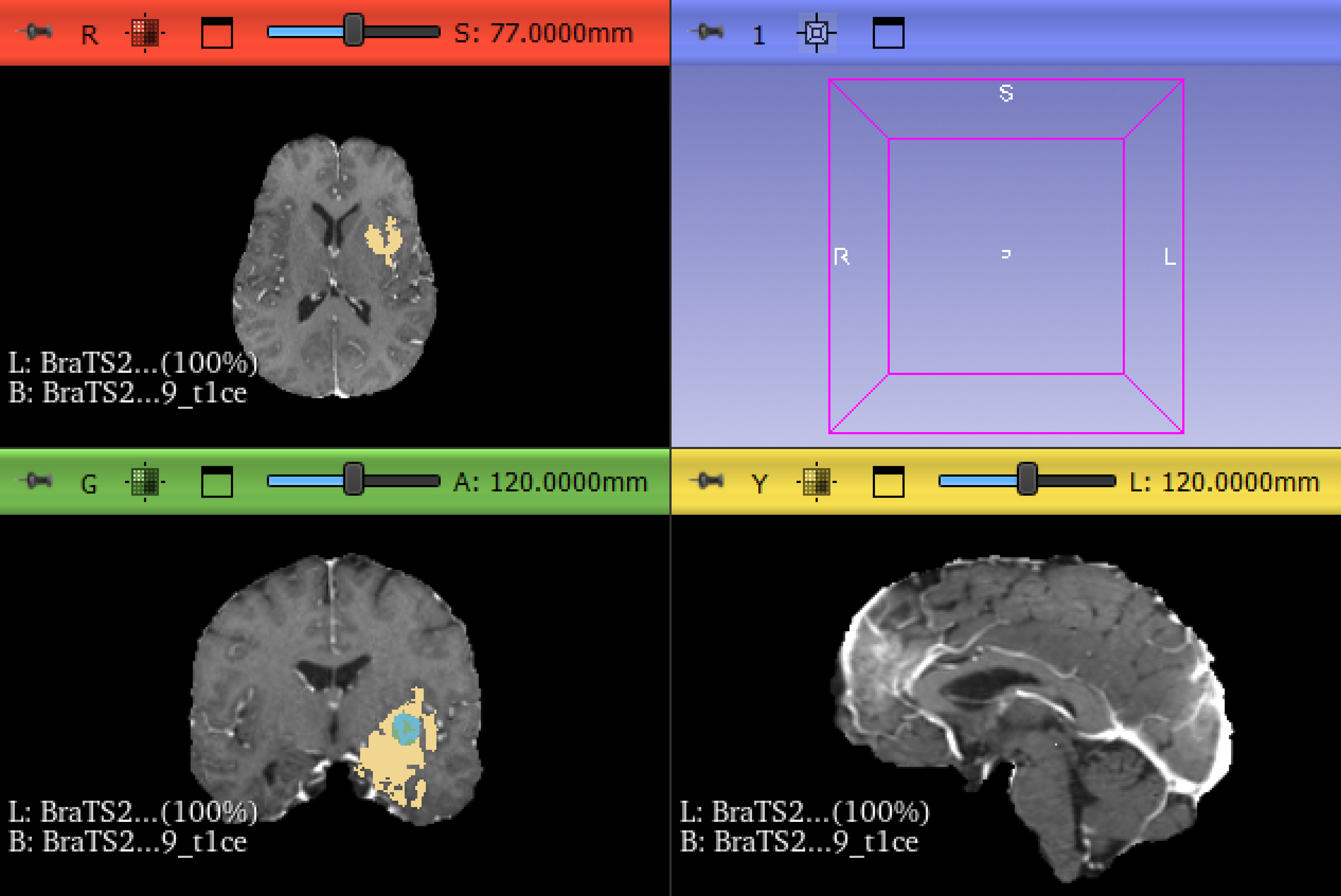}  
  \caption{BraTS20 \texttt{BraTS20\_Training\_059}}
  \label{fig:sub-12-Config2}
\end{subfigure}
\begin{subfigure}{.33\linewidth}
  \centering
  % include second image
  \includegraphics[width=\linewidth]{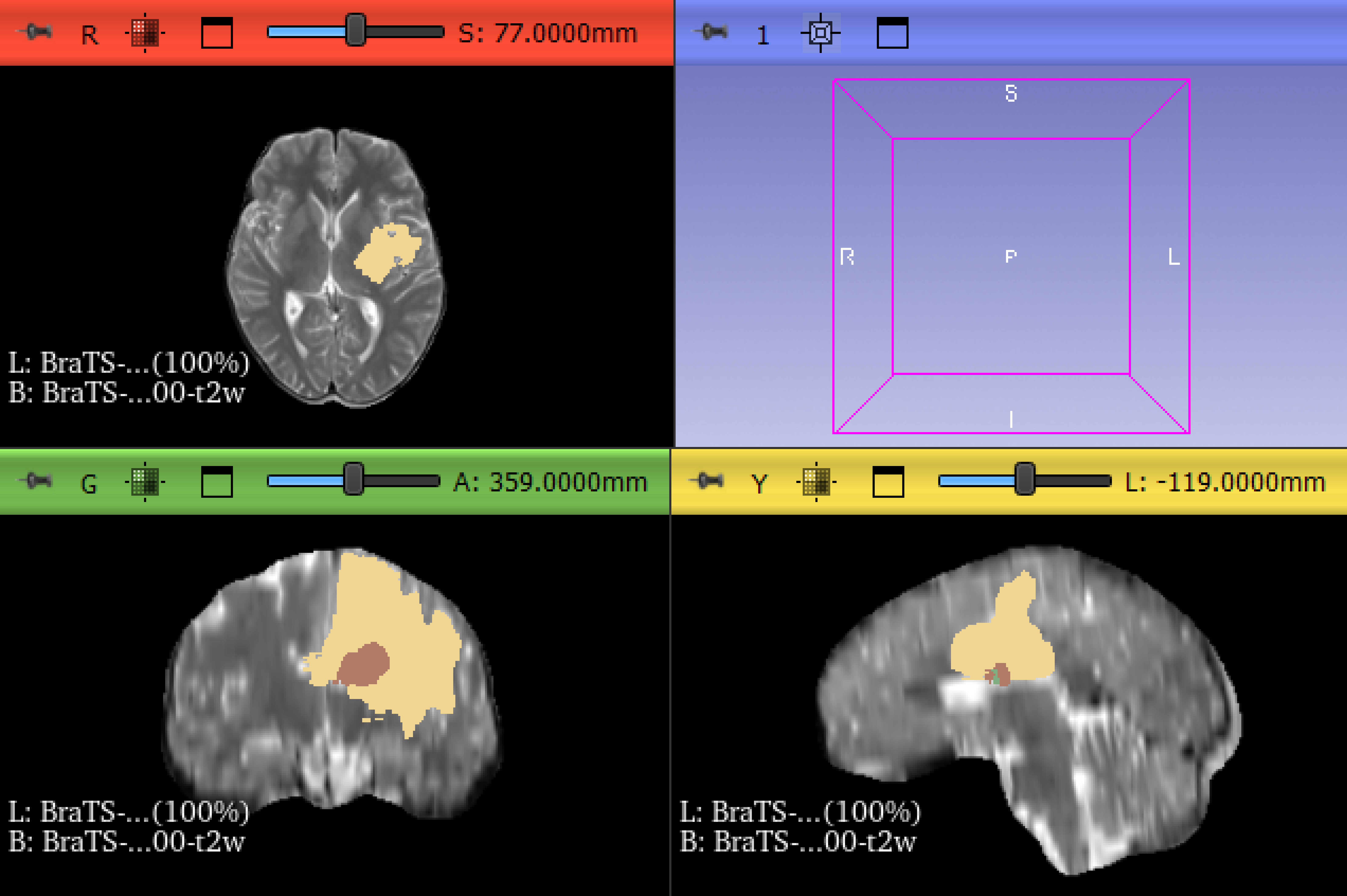}  
  \caption{BraTS-Africa \texttt{BraTS\_SSA\_00050\_000}}
  \label{fig:sub-13-Config2}
\end{subfigure}
% \vspace{-0.25cm}
\caption{Representative BraTS slices illustrating ET sparsity and severe slice-level target-presence imbalance.}
% \vspace{-0.25cm}
\label{Fig:panel_1x3}
\end{figure*}
% --------------------

The Brain Tumor Segmentation (BraTS) benchmark provides standardized, multi-institutional MRI datasets and evaluation protocols for glioma segmentation \cite{bakas2017advancing,menze2018multimodal}. Each subject typically includes T1-weighted, contrast-enhanced T1-weighted (T1ce), T2-weighted, and fluid-attenuated inversion recovery (FLAIR) volumes. These sequences provide complementary tissue information: T1-weighted MRI depicts anatomical structure, T1ce highlights contrast enhancement, and T2-weighted and FLAIR images reveal water-related signal abnormalities, including peritumoral edema \cite{menze2018multimodal,bakas2018identifying}. Their joint use allows segmentation models to combine information about tumor extent and internal composition that is not equally visible in every sequence.

Exploiting this complementary information remains challenging because tumor appearance varies both within and across patients, MRI intensity characteristics depend on scanners and acquisition protocols, and the WT, TC, and ET regions differ substantially in size and prevalence \cite{bakas2018identifying,adewole2025brats}. The model must therefore integrate modality-dependent evidence while preserving small and irregular tumor structures. This difficulty also occurs at the slice level: many two-dimensional (2D) slices contain no ET, whereas target-positive slices may contain only a small enhancing region. As illustrated in \autoref{Fig:panel_1x3}, ET can occupy only a small portion of a tumor volume. Consequently, a positive support example does not imply that the selected region is present in every query slice. A query may contain other tumor components while remaining empty for ET, or it may contain no tumor tissue at all. In both cases, the desired prediction for the absent target is an empty mask.

This motivates the study of target-absent reliability, understood here as the ability to suppress spurious foreground predictions when the selected region is absent. Training predominantly on target-present queries provides limited exposure to this situation and may encourage foreground activation even when the query does not contain the support-defined target. We therefore include target-empty queries during episodic training and examine false-positive activation on ground-truth-empty volume--region pairs alongside the standard segmentation results.

% ----- Data illustration (ET rarity / empty slices) -----
% \begin{table*}[!t]
% \centering
% \setlength{\tabcolsep}{6pt}
% \renewcommand{\arraystretch}{1.0}
% \begin{tabular}{C{0.32\textwidth} C{0.32\textwidth} C{0.32\textwidth}}
% \toprule
% \includegraphics[width=\linewidth,height=4.2cm,keepaspectratio]{Figures/data/b1.png} &
% \includegraphics[width=\linewidth,height=4.2cm,keepaspectratio]{Figures/data/b2.png} &
% \includegraphics[width=\linewidth,height=4.2cm,keepaspectratio]{Figures/data/b3.png} \\
% \bottomrule
% \end{tabular}
% \caption{1$\times$3 panel showing the rarity of ET in BraTS; many slices contain no ET or only small ET regions, causing severe ET positive/negative imbalance.}
% \label{tab:panel_1x3}
% \end{table*}

Deep learning approaches for BraTS have historically been dominated by convolutional encoder--decoder architectures. For example, U-Net \cite{ronneberger2015unet} and its volumetric variants \cite{cciccek20163d,milletari2016v} established strong baselines by combining multi-scale feature extraction with skip connections, while subsequent refinements such as UNet++ \cite{zhou2019unet++} redesigned skip pathways to better exploit multi-resolution features. Self-configuring pipelines such as nnU-Net \cite{isensee2021nnu} further improved robustness through automated design choices and standardized training heuristics. Although convolutional encoder--decoder models provide strong local representations, they may not explicitly model long-range dependencies and broader anatomical context. Transformer-based and hybrid convolutional neural network (CNN)--Transformer architectures have therefore been explored for medical image segmentation \cite{han2022survey}. TransUNet integrates Transformer-based global-context modeling with a U-shaped decoder, whereas Swin-based architectures use shifted-window self-attention to construct hierarchical feature representations \cite{chen2021transunet,liu2021swin,hatamizadeh2021swin,cao2022swin}. In brain tumor MRI segmentation, Swin-PResU combines a Swin Vision Transformer with a pruned residual CNN in a U-Net-style architecture, illustrating how global-context modeling can be coupled with local feature refinement \cite{nguyen2025swin}. These backbone advances motivate contextual representation learning, but they do not directly specify how a labeled support image--mask pair should condition query segmentation or how target-empty queries should be handled in a few-shot setting.

In practical scenarios, dense voxel-wise annotations at scale are expensive, and models must generalize across institutions, populations, and dataset releases. This motivates data-efficient paradigms such as few-shot segmentation (FSS), where a query image is segmented using only a small set of labeled support examples. Prototype-based frameworks have demonstrated strong transferability by summarizing support information into class prototypes and matching them to query features \cite{wang2019panet}. More recent medical few-shot methods have explored mechanisms to mitigate foreground (FG)--background (BG) imbalance and limited supervision, including adaptive prototype pooling, self-supervised training signals \cite{ouyang2020self}, and clinically inspired representations with attention mechanisms \cite{debnath2025fss}. Nevertheless, applying FSS to multi-region brain tumor MRI remains challenging because tumor regions differ markedly in prevalence and appearance and because cross-cohort shifts can alter image characteristics and tumor presentation.

To address these challenges, we propose FSS-UBrain, a multi-region, one-shot FSS framework that performs support-conditioned binary segmentation separately for WT, TC, and ET. For a selected region, the model uses FG--BG representative descriptors derived from a labeled support slice to guide query segmentation through support-guided bottleneck interaction, descriptor-gated decoding, and boundary-guided refinement. During episodic training, positive queries are combined with hard-negative and fully negative queries and an empty-query regularization term to suppress spurious FG activation.

Although FSS-UBrain processes two-dimensional support--query slice pairs, checkpoint selection, threshold optimization, and all final comparisons are performed after reconstructing slice-wise predictions into volumes. We evaluate the method on a held-out BraTS 2020 (BraTS20) test split and under target-supported cross-cohort protocols on BraTS 2023 (BraTS23) and BraTS-Africa. In each external experiment, labeled target-support cases condition inference but are excluded from the query cohort. No target-domain fine-tuning or test-time parameter update is performed. BraTS 2024 (BraTS24) is not included in the quantitative WT/TC/ET analysis because its post-treatment annotation ontology contains a distinct resection-cavity label \cite{de20242024}. BraTS 2017 (BraTS17)--BraTS 2019 (BraTS19) are excluded because duplicate auditing identified overlap with the BraTS20 cohort.
Our main contributions are summarized as follows:
\begin{enumerate}[leftmargin=1.2em]
\setlength{\itemsep}{1pt}
\setlength{\parskip}{1pt}
\vspace{-0.15cm}

\item \textbf{Integrated support-conditioned architecture:} We develop FSS-UBrain, which refines FG and BG descriptors using query context at multiple encoder levels and integrates them into feature adaptation, bottleneck interaction, and decoder reconstruction. Descriptor-gated skip attention, multi-scale descriptor-logit fusion, and boundary-guided refinement form a coordinated prediction pathway for WT, TC, and ET.

\item \textbf{Region-aware negative-query learning:} We combine target-positive, hard-negative, and fully negative queries with empty-query regularization to jointly address target localization and false-positive suppression. Hard-negative queries are defined relative to the selected TC or ET region, distinguishing non-target tumor tissue from fully tumor-free slices.

\item \textbf{Controlled volumetric assessment:} We evaluate reconstructed volumes on held-out BraTS20 and target-supported BraTS23 and BraTS-Africa, keeping support and query cases separate and using validation-only threshold calibration. Component ablations, support-provenance comparisons, negative-query sensitivity analysis, and complexity profiling characterize the framework's performance and trade-offs.

\vspace{-0.15cm}
\end{enumerate}

The remainder of this paper is organized as follows. Section~\ref{Sect:2} reviews related work on brain tumor segmentation, backbone architectures, and few-shot segmentation. Section~\ref{Sect:3} presents the proposed methodology. Section~\ref{sec:experiments} describes the experimental settings and volume-level evaluation protocol. Section~\ref{Sect:5} reports the numerical results and analyses. Section~\ref{sec:discussion} discusses the main findings and limitations. Finally, Section~\ref{sec:conclusion} concludes the paper.

%-----------------------------
\section{Related Work}
\label{Sect:2}
\subsection{Brain Tumor Segmentation and Backbone Architectures}
\label{Sect:2.1}

The BraTS benchmark has become a primary platform for evaluating glioma MRI segmentation, offering standardized multi-institutional data and evaluation protocols for clinically relevant subregions, including WT, TC, and ET, from multimodal MRI \cite{menze2018multimodal,bakas2017advancing,bakas2018identifying}. Despite substantial progress, BraTS-style segmentation remains challenging because of heterogeneous tumor appearance, acquisition variability, and pronounced imbalance across tumor subregions. In particular, ET is often sparse and fragmented. BraTS-Africa further expands BraTS data to include an underrepresented population, making cross-cohort assessment particularly relevant \cite{adewole2025brats}.

Convolutional encoder--decoder networks remain foundational for brain tumor segmentation. U-Net and its volumetric extensions, including 3D U-Net and V-Net, combine multi-scale feature extraction with skip connections, whereas UNet++ improves multi-resolution feature fusion and nnU-Net provides a strong self-configuring training framework \cite{ronneberger2015unet,cciccek20163d,milletari2016v,zhou2019unet++,isensee2021nnu}. These methods establish the CNN backbone family from which later attention-based and Transformer-based designs have evolved.

Transformer-based and hybrid CNN--Transformer architectures have further improved global-context modeling through self-attention \cite{han2022survey}. Representative examples include TransUNet \cite{chen2021transunet}, UNETR \cite{hatamizadeh2022unetr}, SwinUNet \cite{cao2022swin}, and Swin-UNETR \cite{hatamizadeh2021swin}, together with BraTS-oriented variants such as SwinBTS \cite{jiang2022swinbts}, CoTr \cite{xie2021cotr}, nnFormer \cite{zhou2022nnFormer}, and modality-aware cross-attention models \cite{lin2023ckd}. These architectures broaden contextual modeling, although their increased representational capacity must be balanced against computational requirements and the limited availability of annotated medical data. Beyond convolutional and Transformer-based designs, VM-UNet explores visual state-space modeling within a U-shaped architecture for medical image segmentation \cite{ruan2024vm}. Promptable segmentation models, represented by SAM, provide another direction in which prompts guide mask prediction \cite{kirillov2023segment}. These developments broaden the available approaches to contextual representation and target specification. Support-based FSS addresses a related setting in which the target evidence is transferred from an annotated support image to a separate query image, making support representation and support--query interaction central design considerations.

A separate line of neural-network research studies synchronization and control, including applications to image encryption. Examples include passivity-based synchronization under an adaptive event-driven protocol \cite{chandrasekar2026passivity} and sampled-data synchronization with two additive delay components \cite{tamilthendral2026synchronization}. Related work investigates memory-sampled-data control for exponential synchronization of Markovian jump neural networks with mixed delays and partially unknown transition probabilities \cite{radhika2026memory}. Image encryption and decryption concern the protection and recovery of image content, whereas FSS-UBrain predicts tumor-region masks from multimodal MRI using labeled support information.

%----
\subsection{Few-Shot Segmentation}
\label{Sect:2.2}

FSS performs dense prediction on a query image using only a small set of labeled support examples. It is commonly trained through episodic learning, in which support--query episodes encourage the transfer of task-specific information from annotated support images to unlabeled queries \cite{shaban2017one,wang2019panet}. Early approaches transferred support information through conditioning or parameter generation, as exemplified by One-Shot Learning for Semantic Segmentation \cite{shaban2017one}. More recent methods have largely adopted metric- and prototype-based formulations, in which support features and masks are used to construct class representations for dense query matching.

PANet \cite{wang2019panet} is a representative prototype-based method that formulates FSS as a metric-learning problem and introduces prototype alignment regularization to strengthen support--query consistency. Subsequent methods have improved dense correspondence by enriching query features with support-derived priors or by modeling higher-order support--query relationships. For example, the Prior-Guided Feature Enrichment Network \cite{tian2020prior} emphasizes prior-guided feature enrichment, whereas the Hypercorrelation Squeeze Network \cite{min2021hypercorrelation} models multi-level correlations using efficient high-dimensional operators.

Medical FSS introduces additional difficulties, including severe FG--BG imbalance, limited annotated semantic classes, and substantial appearance variation across patients and acquisition settings. ALPNet and SSL-ALPNet \cite{ouyang2020self} address intra-class variation through adaptive local prototype pooling and reduce dependence on manually annotated training classes through self-supervised signals. These studies show the importance of preserving local variation rather than relying exclusively on a single global prototype. Query-aware and cross-attention mechanisms provide complementary strategies for support--query interaction. Q-Net introduces query-informed representations to improve support-conditioned medical image segmentation \cite{shen2023q}, whereas CATNet uses cross-attention to model relationships between support and query features \cite{lin2023few}.

A complementary direction generates multiple support-derived descriptors for each class. GMRD was proposed to address the limited representational capacity of a single prototype by generating multiple representative descriptors from the support image \cite{cheng2024few}. This principle is relevant to heterogeneous tumor MRI, in which tumor-subregion appearance, boundary characteristics, and surrounding BG may vary substantially across cases. FSS-UBrain adopts FG--BG descriptor sets within a distinct support-conditioned architecture that additionally integrates bottleneck interaction, descriptor-gated decoding, and boundary-guided refinement.

Recent medical FSS methods further enrich support representations. SPENet combines a global prototype with an adaptive number of local prototypes and uses query-guided optimal transport to reweight the local support prototypes \cite{fan2026spenet}. BePMI constructs boundary prototypes by clustering inner- and outer-boundary features and uses momentum inference to update prototype representations across adjacent query slices \cite{xu2026bepmi}. These studies demonstrate that local appearance variation, support--query compatibility, and boundary information are established considerations in medical FSS.

\subsection{Research Gaps and Motivation}
\label{Sect:2.3}

The literature reviewed above motivates three considerations for slice-based few-shot brain tumor segmentation. First, heterogeneous tumor appearance requires support representations that preserve variation within the selected region and its surrounding tissue. Multiple descriptors, query-guided prototype enhancement, and boundary-related representations provide established mechanisms for addressing different aspects of this problem \cite{cheng2024few,fan2026spenet,xu2026bepmi}. For the present framework, the architectural question is how to use this support evidence consistently across query-feature adaptation, deep support--query interaction, and decoder-side reconstruction.

Second, sparse subregions such as TC and ET generate many target-empty query slices. At the 2D slice level, a hard-negative query slice is empty for the selected target region while retaining non-empty whole-tumor tissue, whereas a fully negative query slice contains no whole-tumor tissue. Because WT comprises all tumor labels, a WT-empty query is fully negative by construction; hard-negative sampling is therefore relevant only to TC and ET. This distinction is important because strong overlap on target-present queries does not ensure suppression of spurious FG activation when the selected region is absent.

Third, because FSS-UBrain produces slice-wise estimates that are reconstructed into volumes for BraTS-style benchmarking, checkpoint selection, threshold tuning, and final evaluation should be conducted at the reconstructed volume level. Volume-level Dice and 95th-percentile Hausdorff distance (HD95), together with false-positive rates on ground-truth-empty volume--region pairs, cannot be inferred reliably from episode-level slice scores alone. In cross-cohort FSS, the source of support labels, case-level separation between support and query sets, overlap auditing, and the presence or absence of test-time parameter updates should be stated explicitly. When labeled target support cases condition inference without parameter updates, the setting should be described as target-supported FSS rather than zero-shot domain generalization. Together, these considerations guide the design and evaluation of FSS-UBrain.

% Motivated by these gaps, we develop FSS-UBrain as a multi-region one-shot FSS framework that combines FG--BG representative descriptors, support-guided bottleneck interaction, descriptor-gated decoding, and boundary-guided refinement. The training protocol uses positive, hard-negative, and fully negative query slices together with empty-query regularization. The evaluation protocol uses volume-level checkpoint selection and validation-only threshold calibration, followed by case-disjoint target-supported evaluation on BraTS23 and BraTS-Africa.

%-------------------------------
\section{Proposed Methodology}
\label{Sect:3}

This section presents FSS-UBrain, a multi-region, support-conditioned one-shot FSS framework that performs binary segmentation separately for WT, TC, and ET in multimodal brain MRI. Each episode contains one labeled support slice with foreground pixels for the selected region and one query slice; during training, the query may be positive or target-empty. The support image--mask pair provides the FG and BG evidence used to guide query segmentation.

The architecture is designed to address heterogeneous tumor appearance while maintaining selected-region guidance throughout query prediction. Multiple FG and BG descriptors represent support-side appearance variation that may be obscured by a single region average \cite{cheng2024few}, while query-guided refinement makes these representations responsive to the current query. Support-conditioned feature adaptation and bottleneck interaction incorporate target-specific context into query features. Descriptor-gated skip attention and multi-scale descriptor-logit fusion maintain this guidance during spatial reconstruction by combining support-conditioned evidence at different resolutions. Finally, boundary-guided residual refinement uses decoder features and prediction-derived edge cues to correct the preliminary prediction, providing explicit boundary information for delineating irregular tumor contours.

All learnable operations are performed on two-dimensional support--query slice pairs. Deep supervision, descriptor-level supervision, and the alignment loss are applied only during training. During validation, slice-wise predictions are reconstructed into volumes for checkpoint selection and threshold tuning. During testing, the same reconstruction procedure is used for final volume-level evaluation.

\subsection{Problem Formulation}
\label{sec:problem_formulation}

Let \(\mathbf{x}\in\mathbb{R}^{4\times H\times W}\) denote a preprocessed two-dimensional multimodal MRI slice, where \(H\) and \(W\) denote the spatial dimensions. The four channels correspond to FLAIR, T1, T1ce, and T2 modalities. Each slice is associated with a harmonized BraTS-style label map \(\mathbf{y}\in\{0,1,2,4\}^{H\times W}\), where \(y_{ij}\) denotes the label at spatial location \((i,j)\). Under the canonical BraTS20 encoding, labels \(0\), \(1\), \(2\), and \(4\) denote BG, necrotic or non-enhancing tumor core, peritumoral edema, and ET, respectively \cite{bakas2018identifying}.

We consider the clinically relevant tumor-region set \(\mathcal{R}=\{\mathrm{WT},\mathrm{TC},\mathrm{ET}\}\). For each region \(r\in\mathcal{R}\), the multi-class annotation is converted into a binary target mask \(\mathbf{m}^{r}\in\{0,1\}^{H\times W}\). At each spatial location \((i,j)\), the binary target is defined as
\begin{equation}
m_{ij}^{r}
=
\begin{cases}
\mathbb{I}\!\left[y_{ij}\in\{1,2,4\}\right],
& r=\mathrm{WT}, \\[2pt]
\mathbb{I}\!\left[y_{ij}\in\{1,4\}\right],
& r=\mathrm{TC}, \\[2pt]
\mathbb{I}\!\left[y_{ij}=4\right],
& r=\mathrm{ET},
\end{cases}
\label{eq:region_masks}
\end{equation}
where \(\mathbb{I}[\cdot]\) is the indicator function. These definitions imply \(\mathrm{ET}\subseteq\mathrm{TC}\subseteq\mathrm{WT}\). FSS-UBrain predicts the three regions as separate support-conditioned binary tasks rather than as mutually exclusive classes in one multi-class output.

For a selected region \(r\), each one-shot episode contains a labeled support slice \((\mathbf{x}_{s},\mathbf{m}_{s}^{r})\) and a query slice \(\mathbf{x}_{q}\). The corresponding query mask \(\mathbf{m}_{q}^{r}\) is not provided to the model; it is used only to compute training losses and evaluation metrics. Subscripts \(s\) and \(q\) denote the support and query slices, respectively. The support slice is selected such that \(\mathbf{m}_{s}^{r}\) contains FG pixels for the selected region. During episodic training, the query may be positive or target-empty. A hard-negative query is empty for the selected target region while retaining non-empty WT tissue, whereas a fully negative query contains no WT tissue. Because WT comprises all tumor labels, hard-negative queries are applicable only to TC and ET. The construction and sampling ratios of these query types are specified in the experimental protocol. These query types provide complementary training signals. Target-positive queries supervise localization of the selected region. Hard-negative queries require rejection of non-target tumor tissue that may share appearance characteristics with the target, whereas fully negative queries supervise FG suppression in slices without tumor tissue.

Given the support pair \((\mathbf{x}_{s},\mathbf{m}_{s}^{r})\) and query slice \(\mathbf{x}_{q}\), FSS-UBrain produces a dense prediction for the selected region:
\begin{equation}
\begin{aligned}
\mathbf{z}_{q}^{r}
&=
\mathcal{F}_{\theta}\!\left(
\mathbf{x}_{s},
\mathbf{m}_{s}^{r},
\mathbf{x}_{q}
\right),\quad
\mathbf{p}_{q}^{r}
=
\sigma\!\left(\mathbf{z}_{q}^{r}\right),
\end{aligned}
\label{eq:prediction_definition}
\end{equation}
where \(\mathcal{F}_{\theta}\) denotes FSS-UBrain with learnable parameters \(\theta\), \(\mathbf{z}_{q}^{r}\in\mathbb{R}^{1\times H\times W}\) is the query logit map, \(\sigma(\cdot)\) is the element-wise sigmoid function, and \(\mathbf{p}_{q}^{r}\in(0,1)^{1\times H\times W}\) is the predicted FG-probability map. The target region is specified implicitly through the binary support mask rather than through an explicit region-identifier input. In particular, the support mask separates FG and BG support features for representative-descriptor construction, enabling a single parameterized model to adapt its query prediction to WT, TC, or ET. Thresholding, nested-region enforcement, and volume reconstruction are defined later in the evaluation protocol.

For example, in an ET episode, the support mask marks only enhancing tumor as FG, and the resulting FG/BG descriptors guide ET segmentation in a query slice from another patient. In a WT episode, the support mask also includes necrotic/non-enhancing tumor core and peritumoral edema, so tissue treated as BG for ET can become FG for WT. This illustrates how FSS provides region-specific guidance through one labeled support slice per episode, allowing the same trained model to address the selected segmentation task while keeping its learned parameters fixed during inference.

\subsection{Architecture Overview}
\label{sec:architecture_overview}

\begin{figure*}[t!]
\centering
\includegraphics[width=\textwidth]{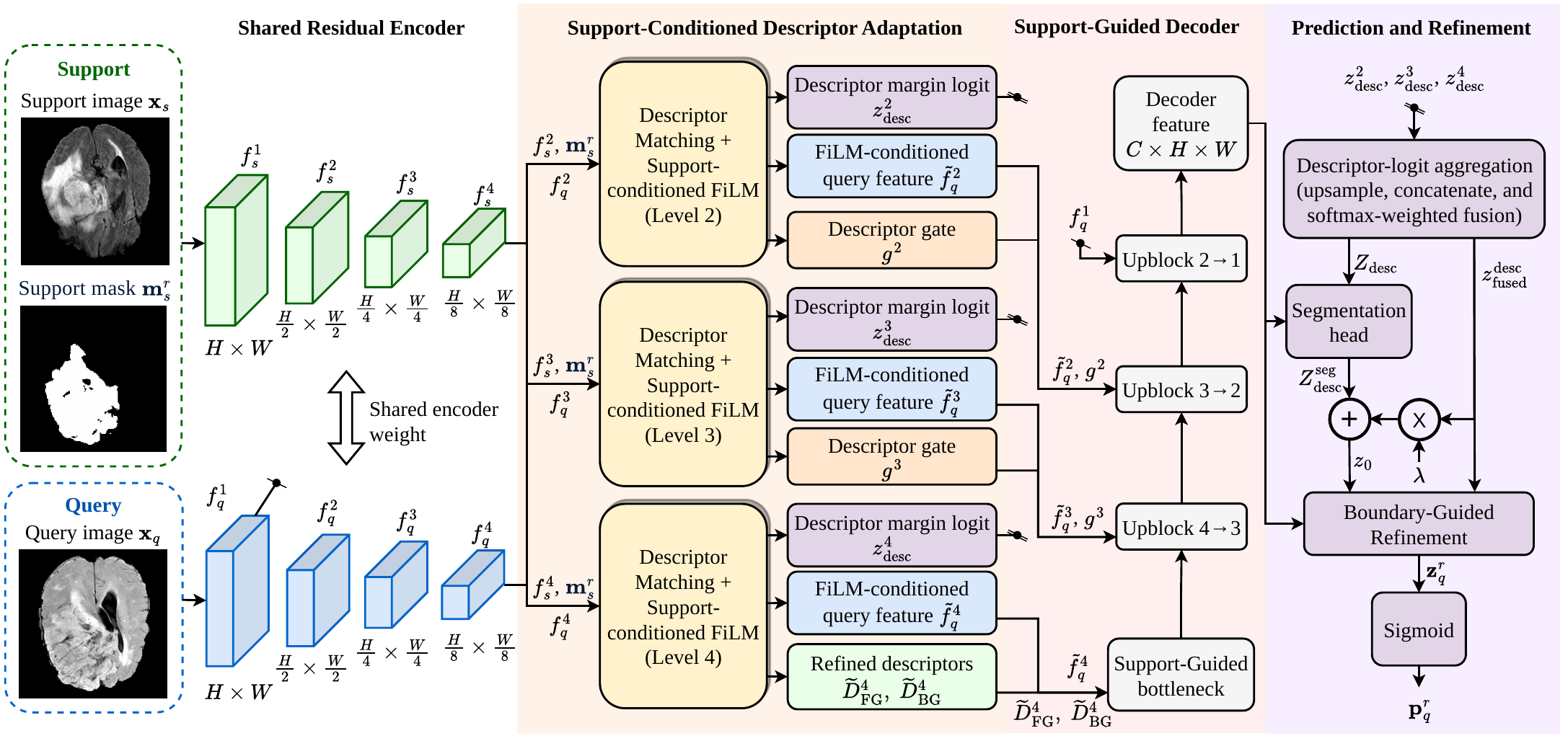}

\vspace{1pt}
\begin{minipage}{\textwidth}
\footnotesize
\textit{Note.} Refined FG/BG descriptors are computed at all levels, but only
\(\widetilde{D}_{\mathrm{FG}}^{4}\) and
\(\widetilde{D}_{\mathrm{BG}}^{4}\)
are explicitly routed to the support-guided bottleneck.
\end{minipage}

\caption{Overview of the proposed FSS-UBrain model architecture.}
\label{fig:fig1}
\end{figure*}
\autoref{fig:fig1} illustrates FSS-UBrain, a support-conditioned one-shot segmentation framework for a selected tumor region \(r\). In this U-shaped encoder--decoder architecture, the encoder extracts hierarchical features at progressively lower spatial resolutions, while the decoder restores resolution for dense prediction. Skip connections transfer higher-resolution encoder features to the decoder, helping combine contextual information with spatial detail \cite{ronneberger2015unet}. In FSS-UBrain, the encoder processes both support and query images, whereas the decoder reconstructs the query mask using support-conditioned query features. The support slice \(\mathbf{x}_s\) and query slice \(\mathbf{x}_q\) are processed by a shared residual encoder, producing hierarchical support and query feature maps \(f_s^\ell\) and \(f_q^\ell\), respectively, at encoder levels \(\ell\in\{1,2,3,4\}\). At level \(\ell\), \( f_s^\ell, f_q^\ell \in \mathbb{R}^{C_\ell\times H_\ell\times W_\ell}, \) where \(C_\ell\), \(H_\ell\), and \(W_\ell\) denote the channel and spatial dimensions. With base channel width \(C=64\), the encoder uses \(C_\ell=2^{\ell-1}C\), \(H_\ell=H/2^{\ell-1}\), and \(W_\ell=W/2^{\ell-1}\). Batch dimensions are omitted throughout this section. Singleton channel dimensions are retained when they clarify single-channel descriptor logits or gates.

Support-conditioned descriptor adaptation is performed at levels \(\ell\in\{2,3,4\}\). At each level, the support mask \(\mathbf{m}_s^r\) separates support features into FG and BG regions for descriptor construction. The resulting descriptor sets condition the corresponding query feature and produce a Feature-wise Linear Modulation (FiLM)-conditioned query feature \(\tilde{f}_q^\ell\), a descriptor margin logit \(z_{\mathrm{desc}}^\ell\in \mathbb{R}^{1\times H_\ell\times W_\ell}\), and a descriptor gate \(g^\ell\in(0,1)^{1\times H_\ell\times W_\ell}\). The refined level-4 descriptor sets \(\widetilde{D}_{\mathrm{FG}}^4\) and \(\widetilde{D}_{\mathrm{BG}}^4\) are additionally supplied to the support-guided bottleneck. The decoder subsequently reconstructs the query prediction through descriptor-gated skip attention, multi-scale descriptor-logit fusion, and boundary-guided refinement.

\subsubsection{Shared Residual Encoder}
\label{sec:shared_residual_encoder}

The support and query branches share the same encoder parameters so that both inputs are represented through a common learned feature transformation. This is useful for support-conditioned segmentation because the subsequent descriptor-matching operations compare support-derived representations with query features. Weight sharing provides a consistent feature mapping for these comparisons and avoids maintaining separate sets of encoder parameters for the two branches.

The encoder begins with a \(3\times3\) convolution, Group Normalization, and Sigmoid Linear Unit (SiLU) activation, followed by a residual refinement block. Three downsampling stages progressively reduce spatial resolution while increasing channel capacity. Each stage contains a stride-two convolutional transformation followed by residual feature refinement. Residual connections allow a block to learn a feature correction while preserving a direct information path, thereby facilitating feature and gradient propagation through the encoder \cite{he2016deep}. This supports the extraction of hierarchical representations needed for both local tumor detail and broader anatomical context. Group Normalization is used because it does not depend on batch-level statistics and is suitable for the small mini-batches used in episodic training \cite{wu2018group}.

The full-resolution features \(f_s^1\) and \(f_q^1\) preserve local spatial detail for the final decoder stage. Levels \(2\), \(3\), and \(4\) provide progressively more abstract representations at one-half, one-quarter, and one-eighth of the input resolution, respectively. At level \(4\), a pyramid-context module applies parallel \(1\times1\) convolution and dilated \(3\times3\) convolution branches with dilation rates of \(2\) and \(4\). Their outputs are concatenated and projected back to the original channel dimension. This operation enriches the deepest representation with multi-scale contextual information, following the general use of dilated convolutions for context aggregation in semantic segmentation \cite{chen2017rethinking}.

\subsubsection{Support-Conditioned Descriptor Adaptation}
\label{sec:descriptor_adaptation}

\begin{figure*}[t!]
\centering
\includegraphics[width=\textwidth]{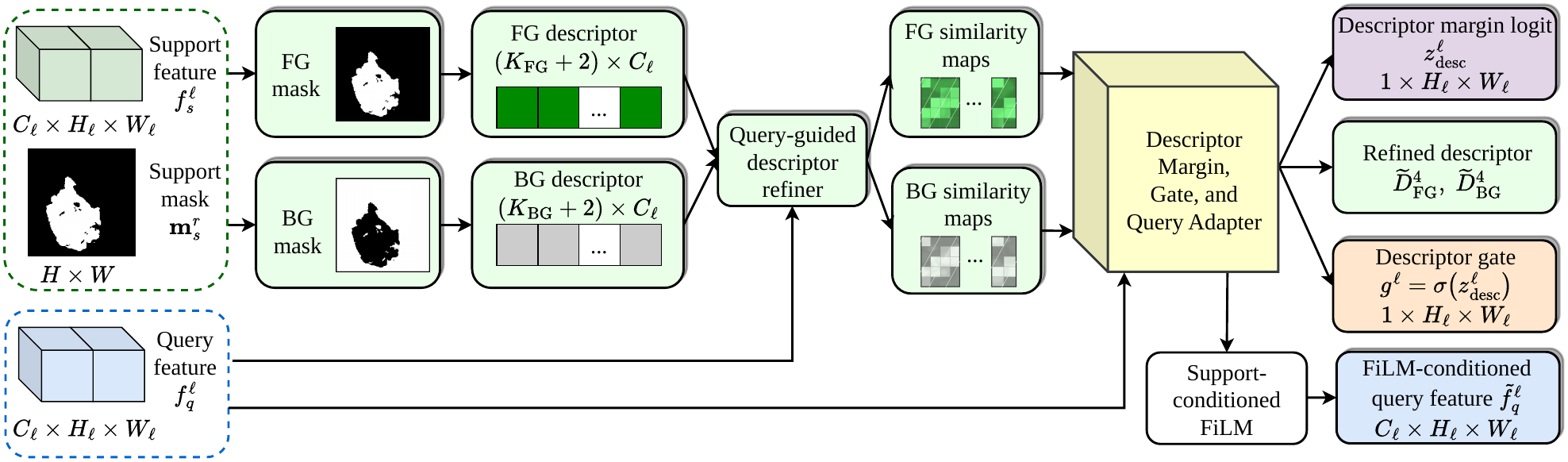}
\caption{Support-conditioned descriptor adaptation at encoder level \(\ell\). Masked support features form FG and BG descriptors that are query-refined, matched to query features, and used for descriptor gating and FiLM modulation.}
\label{fig:descriptor_adaptation}
\end{figure*}

\autoref{fig:descriptor_adaptation} presents the support-conditioned descriptor-adaptation block at encoder level \(\ell\). The block receives the support feature \(f_s^\ell\), the support mask \(\mathbf{m}_s^r\), and the query feature \(f_q^\ell\). It constructs multiple FG and BG descriptors from the support representation, refines these descriptors using query context, and matches them densely against the query feature. The resulting support-conditioned evidence is used to generate \(z_{\mathrm{desc}}^\ell\), \(g^\ell\), and \(\tilde{f}_q^\ell\).

For each matching level, \(\mathbf m_s^r\) is resized to \(H_\ell\times W_\ell\) by nearest-neighbor interpolation and dilated by one feature-grid pixel. We denote the resulting working mask by \(\bar{\mathbf m}_{s,\ell}^{r}\); its complement \(\mathbf 1-\bar{\mathbf m}_{s,\ell}^{r}\) defines the BG region. For a feature-aligned mask \(\mathbf{a}\in[0,1]^{H_\ell\times W_\ell}\), we obtain a support-derived region summary through masked average pooling (MAP), consistent with the prototype-estimation paradigm in few-shot segmentation \cite{wang2019panet}:
\begin{equation}
\operatorname{MAP}(f_s^\ell,\mathbf{a})
=
\frac{
\sum_{i=1}^{H_\ell}\sum_{j=1}^{W_\ell}
[f_s^\ell]_{:,i,j}[\mathbf{a}]_{ij}
}{
\max\!\left(
\epsilon_1,
\sum_{i=1}^{H_\ell}\sum_{j=1}^{W_\ell}
[\mathbf{a}]_{ij}
\right)
},
% \quad
% \epsilon=10^{-6},
\label{eq:masked_average_pooling}
\end{equation}
where \([f_s^\ell]_{:,i,j}\in\mathbb{R}^{C_\ell}\) denotes the support-feature vector at spatial location \((i,j)\), and \(\epsilon_1 =10^{-6}\) is a numerical-stability constant that prevents division by zero when the corresponding region contains no pixels on the feature grid.

The FG descriptor set is denoted by \( D_{\mathrm{FG}}^\ell \in \mathbb{R}^{(K_{\mathrm{FG}}+2)\times C_\ell}, \) where \(K_{\mathrm{FG}}\) denotes the number of primary FG descriptors generated by a learnable descriptor projection. The remaining two descriptors are support-derived: a masked-average FG descriptor and an inner-boundary descriptor. The latter is obtained from the difference between the working FG mask and its eroded version.

Similarly, the BG descriptor set is denoted by \( D_{\mathrm{BG}}^\ell \in \mathbb{R}^{(K_{\mathrm{BG}}+2)\times C_\ell}, \) where \(K_{\mathrm{BG}}\) denotes the number of primary BG descriptors. The remaining BG descriptors are a global BG descriptor and an outer-boundary descriptor. The outer boundary is defined as the one-pixel ring surrounding the working FG region. This multiple-descriptor formulation is motivated by representative-descriptor learning, which preserves support-side appearance variation beyond a single global prototype \cite{cheng2024few}.

The primary descriptors are generated from dense masked feature canvases. In the FG branch, feature vectors outside the working FG region are replaced by the masked-average FG feature. The resulting canvas is adaptively pooled and processed by a learnable descriptor projection to produce \(K_{\mathrm{FG}}\) primary descriptors. The BG branch is constructed analogously using the complementary support region. This design preserves local support-side variation while maintaining class-specific descriptor generation.

The initial descriptor sets are refined using query context. For each descriptor, normalized descriptor--query similarities are computed over the spatial query grid. A spatial softmax produces a descriptor-specific query-context vector. A LayerNorm--multilayer perceptron (MLP) update then predicts a descriptor correction from the difference between the query context and the initial descriptor. This correction is incorporated through a bounded learned blend, producing the refined descriptor sets \(\widetilde{D}_{\mathrm{FG}}^\ell\) and \(\widetilde{D}_{\mathrm{BG}}^\ell\). The refinement preserves descriptor cardinality, such that \( \widetilde{D}_{\mathrm{FG}}^\ell \in \mathbb{R}^{(K_{\mathrm{FG}}+2)\times C_\ell}
\quad\text{and}\quad \widetilde{D}_{\mathrm{BG}}^\ell \in \mathbb{R}^{(K_{\mathrm{BG}}+2)\times C_\ell}. \) The refined descriptors are matched densely against channel-normalized query features to form descriptor-wise similarity maps. Separate convolutional heads transform the FG and BG similarity maps into single-channel spatial logits. Their difference forms the descriptor margin logit \(z_{\mathrm{desc}}^\ell\), while its sigmoid transform defines the descriptor gate \(g^\ell\). Descriptor-wise softmax weights also produce FG and BG context features together with their affinity maps. The query adapter receives the original query feature, the FG and BG context features, the two affinity maps, and a scaled and clipped descriptor margin. It predicts a residual correction to \(f_q^\ell\). The associated scalar gain is initialized at zero, so the adapter initially preserves the unconditioned query feature.

The refined descriptor sets further condition the query feature through FiLM-style feature-wise affine modulation \cite{perez2018film}. For a descriptor-adapted feature map \(h\in\mathbb{R}^{C_\ell\times H_\ell\times W_\ell}\), the implementation adopts the following identity-centered bounded parameterization:
\begin{equation}
\operatorname{FiLM}(h;\gamma,\beta)
=
h\odot\left(1+\rho\tanh(\gamma)\right)
+
\rho\tanh(\beta),
\quad
\rho=0.15,
\label{eq:support_film}
\end{equation}
where \(\gamma,\beta\in\mathbb{R}^{C_\ell\times1\times1}\) are channel-wise modulation parameters, and \(\odot\) denotes element-wise multiplication. They are predicted by an MLP from the LayerNorm-normalized concatenation of the mean vectors of the refined FG and BG descriptor sets, \(\widetilde{D}_{\mathrm{FG}}^\ell\) and \(\widetilde{D}_{\mathrm{BG}}^\ell\), respectively. The final MLP layer is initialized to produce zero modulation parameters, such that the transformation is initialized as the identity mapping, while the bounded hyperbolic-tangent parameterization limits subsequent descriptor-induced feature perturbations. The resulting feature is denoted by \(\tilde{f}_q^\ell\).

\subsubsection{Support-Guided Decoder and Prediction Refinement}
\label{sec:support_guided_decoder}

At level \(4\), the FiLM-conditioned feature \(\tilde{f}_q^4\) and the refined descriptor sets \(\widetilde{D}_{\mathrm{FG}}^4\) and \(\widetilde{D}_{\mathrm{BG}}^4\) are processed by the support-guided bottleneck. The spatial locations of \(\tilde{f}_q^4\) are flattened into query tokens. The concatenated FG and BG descriptors serve as memory tokens, while an additional support token is generated from the mean FG and BG descriptors. Multi-head attention retrieves support-relevant descriptor information for each query token, followed by a feed-forward update and a residual projection back to the original feature space. This mechanism follows the standard query--key--value attention principle \cite{vaswani2017attention}.

The decoder restores spatial resolution through three upsampling blocks. The support-guided bottleneck output is fused with \(\tilde{f}_q^3\) to produce \(x_3\). The second block fuses \(x_3\) with \(\tilde{f}_q^2\) to produce \(x_2\). The final upsampling block fuses \(x_2\) with the full-resolution query feature \(f_q^1\) to produce the final full-resolution decoder feature \(x_1\). The channel dimensions of \(x_3\), \(x_2\), and \(x_1\) are \(C_3\), \(C_2\), and \(C_1\), respectively.

Each upsampling block first applies a transposed convolution to increase the spatial resolution of the decoder feature. The corresponding query-side skip feature is then filtered through additive attention before concatenation and residual convolutional refinement. Let \(u^\ell\) denote the spatially aligned upsampled decoder feature and \(s^\ell\) denote the corresponding query-side skip feature. Specifically, \(u^3\) is obtained from the support-guided bottleneck output, \(u^2\) from \(x_3\), and \(u^1\) from \(x_2\), whereas \(s^3=\tilde{f}_q^3\), \(s^2=\tilde{f}_q^2\), and \(s^1=f_q^1\). We use an additive attention gate inspired by Attention U-Net \cite{oktay2018attention}:
\begin{equation}
a^\ell
=
\sigma\!\left(
\psi\!\left(
\phi\!\left(
\operatorname{GN}(W_g u^\ell)
+
\operatorname{GN}(W_s s^\ell)
\right)
\right)
\right),
\quad
\widehat{s}^\ell
=
s^\ell\odot a^\ell,
\label{eq:additive_attention_gate}
\end{equation}
where \(W_g\) and \(W_s\) denote learned \(1\times1\) projections, \(\operatorname{GN}(\cdot)\) denotes Group Normalization, \(\phi(\cdot)\) is the SiLU nonlinearity, \(\psi(\cdot)\) maps the fused representation to a single-channel attention score, and \(\odot\) denotes element-wise multiplication.

At levels \(3\) and \(2\), the descriptor gates \(g^3\) and \(g^2\) further modulate the corresponding skip-attention maps. After spatial alignment, each gate is clipped to the probability range and rescales the base attention response by \(1+0.35(g^\ell-0.5)\). No descriptor gate is applied to the final skip fusion with \(f_q^1\). Although \(g^4\) is computed at level \(4\), it has no corresponding encoder--decoder skip connection and is therefore not used by an upsampling block.

The descriptor margin logits from levels \(2\), \(3\), and \(4\) are bilinearly upsampled to the spatial resolution of \(x_1\). Their channel-wise concatenation is denoted by \( Z_{\mathrm{desc}} \in \mathbb{R}^{3\times H\times W}. \) This tensor is concatenated with \(x_1\) and processed by the segmentation head, producing $Z_{\textrm{desc}}^{\textrm{seg}} \in \mathbb{R}^{1\times H\times W}$. In parallel, the three descriptor logits are fused using learned softmax-normalized scale weights \(\alpha_2,\alpha_3,\alpha_4\), producing \( z_{\mathrm{fused}}^{\mathrm{desc}} \in \mathbb{R}^{1\times H\times W}. \) The segmentation-head logit receives an additional residual contribution from \(z_{\mathrm{fused}}^{\mathrm{desc}}\), scaled by a learned sigmoid-bounded scalar \(\lambda\in(0,1)\). The resulting preliminary logit is denoted by \( z_0 = Z_{\textrm{desc}}^{\textrm{seg}} + \lambda z_{\mathrm{fused}}^{\mathrm{desc}} \in \mathbb{R}^{1\times H\times W}. \)
% \textcolor{red}{Provide the explicit form of $z_0$}

Boundary-guided refinement receives the final full-resolution decoder feature \(x_1\), the preliminary logit \(z_0\), and the fused descriptor logit \(z_{\mathrm{fused}}^{\mathrm{desc}}\). It derives detached FG probability maps from the two logits and computes their soft edge maps using a \(3\times3\) morphological gradient based on dilation and erosion. The feature \(x_1\), the two probability maps, and the two edge maps are concatenated and processed by a residual convolutional trunk followed by a residual-logit head to produce \(\Delta_{\mathrm{ref}}\in\mathbb{R}^{1\times H\times W}\). The final query logit map is defined as
\begin{equation}
\mathbf{z}_q^r
=
z_0
+
\rho_{\mathrm{ref}}
\tanh\!\left(\alpha_{\mathrm{ref}}\right)
\Delta_{\mathrm{ref}},
\quad
\rho_{\mathrm{ref}}=0.35,
\label{eq:boundary_refinement}
\end{equation}
where \(\Delta_{\mathrm{ref}}\in\mathbb{R}^{1\times H\times W}\) is the residual logit correction produced by the residual-logit head, \(\alpha_{\mathrm{ref}}\in\mathbb{R}\) is a learned scalar residual-gain parameter, and \(\rho_{\mathrm{ref}}=0.35\) specifies the maximum magnitude of the scalar gain \(\rho_{\mathrm{ref}}\tanh(\alpha_{\mathrm{ref}})\). Here, \(\tanh(\cdot)\) denotes the hyperbolic tangent function. The residual-gain parameter is initialized to \(1\), whereas the residual-logit head is zero-initialized. Consequently, at initialization, \(\Delta_{\mathrm{ref}}=0\), and the refinement block preserves the preliminary logit \(z_0\). The nonzero residual gain allows the residual-logit head to receive gradients during training; once its weights depart from zero, gradients also propagate through the preceding refinement trunk. The corresponding probability map \(\mathbf{p}_q^r\) is obtained using the sigmoid operation defined in Section~\ref{sec:problem_formulation}. Although the implementation also returns an auxiliary boundary-logit output, no separate supervision or inference decision is based on that output.

\subsection{Training Objective}
\label{sec:training_objective}

The primary supervision is applied to query-side outputs. During training, the final refined logit \(\mathbf{z}_q^r\) is accompanied by three auxiliary logits derived from \(x_1\), \(x_2\), and \(x_3\). The auxiliary logits from \(x_2\) and \(x_3\) are bilinearly upsampled to the final output resolution, whereas the auxiliary logit from \(x_1\) is already defined at that resolution. Consequently, all deep-supervision outputs are evaluated against the full-resolution query mask \(\mathbf{m}_q^r\).

For a supervised output, let \(p_{ij}\in(0,1)\) denote the predicted FG probability at pixel \((i,j)\), and let \(m_{ij}\in\{0,1\}\) denote the corresponding binary target value. We use the following smoothed Dice objective, based on the Dice formulation widely adopted in medical image segmentation \cite{milletari2016v}:
\begin{equation}
\mathcal{L}_{\mathrm{Dice}}
=
1-
\frac{
2\sum_{(i,j)\in\Omega}p_{ij}m_{ij} + \epsilon_2
}{
\sum_{(i,j)\in\Omega}p_{ij}^2
+
\sum_{(i,j)\in\Omega}m_{ij}^2
+
\epsilon_2
},
% \qquad
% \epsilon=10^{-6},
\label{eq:dice_loss}
\end{equation}
where \(\Omega\) denotes the spatial pixel set of the supervised output and $\epsilon_2 = 10^{-6}$ is a small constant used for numerical stability. The loss is computed independently for each sample in a mini-batch and then averaged over the batch.

For each supervised decoder output, the segmentation loss combines the smoothed Dice loss in Eq.~\eqref{eq:dice_loss}, boundary-weighted binary cross-entropy (BCE), and focal binary cross-entropy with coefficients \(0.45\), \(0.45\), and \(0.10\), respectively. For the boundary-weighted BCE term, a target-derived edge map is computed using \(3\times3\) morphological dilation and erosion. The pixel-wise BCE weight is one plus \(0.20\) times the edge value, thereby emphasizing boundary errors without introducing a separately supervised boundary output. The focal term uses focusing exponent \(\gamma_{\mathrm{foc}}=2\) \cite{lin2017focal}. When necessary, auxiliary logits are bilinearly upsampled to the final output resolution before supervision by the full-resolution query mask.

For target-empty queries, an empty-query regularization term is added to each supervised decoder output to explicitly discourage foreground activation. It penalizes the mean FG probability together with \(0.25\) times the maximum FG probability. The mean-probability component discourages spatially distributed activation, whereas the maximum-probability component penalizes a strong local response that may remain small after averaging over the full image.

An additional hinge penalty is activated when the mean FG probability exceeds \(0.035\) for a hard-negative query or \(0.015\) for a fully negative query. The hinge contribution has relative weight \(0.10\), and the complete empty-query term is scaled by \(0.16\). These hinge thresholds determine when the additional penalty becomes active; the reference mask for the selected region remains empty in both cases. The term is applied only to target-empty queries, while target-positive queries retain the foreground supervision provided by the segmentation loss. Together, negative-query sampling and this regularization provide explicit learning signals for rejecting non-target tissue and suppressing spurious FG responses.

Let \(d=0\) denote the final refined logit \(\mathbf{z}_q^r\), and let \(d\in\{1,2,3\}\) denote the auxiliary logits derived from \(x_1\), \(x_2\), and \(x_3\), respectively. For output \(d\), \(\mathcal{L}_{\mathrm{seg}}^{(d)}\) denotes the three-term segmentation loss described above, and \(\mathcal{L}_{\mathrm{empty}}^{(d)}\) denotes the corresponding empty-query regularization term. The latter is evaluated only for target-empty samples and is zero for target-positive samples. Following deep supervision \cite{lee2015deeply}, the decoder objective is
\begin{equation}
\begin{aligned}
\mathcal{L}_{\mathrm{deep}}
&=
\sum\nolimits_{d=0}^{3} w_d
\left(
\mathcal{L}_{\mathrm{seg}}^{(d)}
+
\mathcal{L}_{\mathrm{empty}}^{(d)}
\right),
% \\
% w_0&=0.62,\qquad w_1=0.20,\\
% w_2&=0.12,\qquad w_3=0.06.
\end{aligned}
\label{eq:deep_supervised_loss}
\end{equation}
where \(w_d\) denotes the supervision weight assigned to output \(d\), which is chosen based on empirical experiments. Here, \(w_0 = 0.62\) corresponds to the final refined output, whereas \(w_1 = 0.2\), \(w_2 = 0.12\), and \(w_3 = 0.06\) correspond to the auxiliary outputs derived from \(x_1\), \(x_2\), and \(x_3\), respectively. Descriptor-level supervision is applied to \(z_{\mathrm{desc}}^2\), \(z_{\mathrm{desc}}^3\), \(z_{\mathrm{desc}}^4\), and \(z_{\mathrm{fused}}^{\mathrm{desc}}\). Each descriptor logit receives the same three-term segmentation loss after the query mask is resized to the corresponding spatial resolution. Their mean defines \(\mathcal{L}_{\mathrm{desc}}\). Empty-query regularization is not applied to this auxiliary descriptor supervision.

A reverse support--query alignment term is applied only to target-positive episodes. A detached probability map derived from the final query prediction is resized to encoder level \(4\) and used to construct query-derived FG and BG descriptors. These descriptors are matched to the level-4 support feature, and the resulting support-side margin logit is supervised using the resized support mask. The resulting loss, denoted by \(\mathcal{L}_{\mathrm{align}}\), follows the prototype-alignment principle of PANet \cite{wang2019panet}. It is set to zero when a mini-batch contains no target-positive sample.

The complete training objective is
\begin{equation}
\mathcal{L}_{\mathrm{total}}
=
\mathcal{L}_{\mathrm{deep}}
+
\delta_1\,\mathcal{L}_{\mathrm{desc}}
+
\delta_2\,\mathcal{L}_{\mathrm{align}},
\label{eq:total_training_objective}
\end{equation}
where $\delta_1 = 0.08$ and $\delta_2 = 0.15$ are the loss weights, chosen based on empirical experiments and task-specific evaluations. All loss terms and auxiliary outputs are used only during training. During inference, FSS-UBrain uses only the final query logit \(\mathbf{z}_q^r\) and probability map \(\mathbf{p}_q^r\).

% =========================================================
% Section 4: Experimental Details
% =========================================================
\section{Experimental Details}
\label{sec:experiments}

\subsection{Datasets, Label Harmonization, and Data Processing}
\label{sec:data_preparation}

\subsubsection{Source Cohort and Held-Out Split}

BraTS20 was used as the source cohort for model development \cite{brats2020data,bakas2018identifying}. Its 369 labeled cases were partitioned once at the case level using seed \(1337\). First, \(10\%\) of the complete cohort was reserved for held-out testing, yielding 37 test cases. Next, \(10\%\) of the remaining 332 cases was assigned to validation, yielding 33 validation cases and leaving 299 cases for training. The held-out test partition was not used for checkpoint selection, threshold calibration, architecture selection, or hyperparameter adjustment.

All compared models were trained on the same BraTS20 training partition. The validation partition was used for model selection and calibration, whereas the held-out partition was used only for final in-domain evaluation.

\subsubsection{External Cohorts and Overlap Audit}

External evaluation used the publicly labeled BraTS23 adult-glioma training cohort and the glioma subset of BraTS-Africa. These datasets permit harmonization with the BraTS20 WT, TC, and ET composite regions. The BraTS24 adult-glioma (GLI) cohort was not included in the quantitative cross-cohort analysis because it comprises post-treatment MRI and introduces a distinct resection-cavity label \cite{de20242024}. Merging this label with ET would alter the clinical meaning of ET and its composite regions. These cohorts assess robustness beyond the BraTS20 source cohort under cross-release variation and, in the case of BraTS-Africa, population shift \cite{brats2023portal,adewole2025brats}.

Before external evaluation, each candidate case was verified to contain all four MRI modalities and a spatially compatible segmentation mask. Potential overlap was audited against the complete BraTS20 cohort, including its training, validation, and held-out test partitions. The audit compared canonicalized case identifiers and full content fingerprints computed from the FLAIR, T1, T1ce, T2, and segmentation arrays. A target case was excluded whenever either criterion indicated an overlap with BraTS20.

After this audit, BraTS17 and BraTS18 contained no retained non-overlapping cases. BraTS19 retained only one case, \texttt{BraTS19\_CBICA\_BLJ\_1}, which is insufficient for a meaningful cohort-level analysis. Accordingly, BraTS17--BraTS19 are not reported as independent external cohorts because duplicate auditing identified overlap with the complete BraTS20 cohort.

\subsubsection{Label Harmonization}

All datasets were mapped to the common WT, TC, and ET region definitions introduced in Section~\ref{sec:problem_formulation}. Under the canonical BraTS20 encoding, labels 1, 2, and 4 represent necrotic or non-enhancing tumor core, edema, and enhancing tumor, respectively. BraTS20 encodes enhancing tumor as label 4, whereas the retained external cohorts use their release-specific enhancing-tumor encoding. Before constructing WT, TC, and ET, each retained cohort was interpreted using its own label map. Whole tumor comprises all non-BG tumor components, tumor core comprises the necrotic or non-enhancing component together with enhancing tumor, and enhancing tumor comprises the enhancing component only. This harmonization does not imply that labels 3 and 4 coexist within an individual release; rather, it provides a consistent clinical interpretation of WT, TC, and ET across datasets with different enhancing-tumor encodings.

\subsubsection{Slice-Level Preprocessing and Augmentation}

Each axial slice was represented by four channels in the order FLAIR, T1, T1ce, and T2. For each case and modality, normalization statistics were computed from nonzero voxels over the entire volume. Nonzero intensities were standardized using these statistics, whereas zero-valued BG voxels remained zero. The same preprocessing was applied to support and query inputs.

\autoref{tab:preprocessing_augmentation} summarizes the preprocessing and training-time augmentation pipeline. Geometric transformations were applied jointly to each image--mask pair to preserve spatial alignment. Support and query pairs were augmented independently during training; validation and test inputs were not augmented.

\begin{table*}[ht!]
\centering
\caption{Preprocessing and training-time augmentation applied to support and query inputs.}
\label{tab:preprocessing_augmentation}
\renewcommand{\arraystretch}{1.08}
\footnotesize
% \resizebox{0.95\textwidth}{!}{%
\begin{tabular}{|p{10.2cm}|p{4.5cm}|p{2.2cm}|}
\hline
\centering\textbf{Preprocessing or Augmentation Step and Procedure}
& \centering\textbf{Purpose}\arraybackslash
& \centering\textbf{Applied During}\arraybackslash \\
\hline\hline

\textbf{Modality organization.} Each axial input is formed by stacking FLAIR, T1, T1ce, and T2 as a four-channel slice.
& Preserves the standard multimodal BraTS input representation.
& Training, validation, and testing \\
\hline

\textbf{Region-mask construction.} Binary WT, TC, and ET masks are generated using the unified region definitions described in Section~\ref{sec:problem_formulation}.
& Ensures a consistent clinical-region interpretation across datasets.
& Training, validation, and testing \\
\hline

\textbf{Per-case normalization.} For each modality, the mean and standard deviation are computed from nonzero voxels across the complete case volume. Only nonzero voxels are standardized, while zero-valued BG remains zero.
& Reduces inter-case intensity variation without altering the zero-valued BG convention.
& Training, validation, and testing \\
\hline

\textbf{Spatial resizing.} Images are resized to \(256\times256\) using bilinear interpolation; binary masks are resized using nearest-neighbor interpolation.
& Provides a fixed input size while avoiding interpolation-induced label mixing.
& Training, validation, and testing \\
\hline

\textbf{Horizontal and vertical flips.} Each image--mask pair is independently flipped horizontally and vertically with probability \(0.5\).
& Increases invariance to in-plane orientation variation.
& Training only \\
\hline

\textbf{Right-angle rotation.} With probability \(0.5\), an image--mask pair is rotated by \(0^\circ\), \(90^\circ\), \(180^\circ\), or \(270^\circ\).
& Improves robustness to in-plane orientation variation while preserving voxel-grid alignment.
& Training only \\
\hline

\textbf{Random affine transformation.} With probability \(0.35\), an affine transformation is applied using a rotation angle sampled from \([-8^\circ,8^\circ]\), a scale factor sampled from \([0.95,1.05]\),
and translations of up to \(4\%\) of the image size along both spatial axes. Images use bilinear interpolation and masks use nearest-neighbor interpolation.
& Increases tolerance to modest geometric variation while preserving image--mask correspondence.
& Training only \\
\hline

\textbf{Intensity perturbation.} With probability \(0.5\), image intensities are multiplicatively scaled by a factor from \([0.90,1.10]\) and additively shifted by a value from \([-0.05,0.05]\).
& Improves robustness to moderate intensity-scale and brightness variation.
& Training only \\
\hline

\textbf{Numerical stabilization.} Non-finite image values are replaced with zero before further processing. During augmentation, image intensities are clipped to \([-8,8]\) after the geometric and intensity transformations.
& Prevents invalid numerical values and limits extreme augmented intensities.
& All stages; clipping during training only \\
\hline
\end{tabular}%
% }
\end{table*}

\subsection{One-Shot Training Configuration}
\label{sec:implementation_details}

FSS-UBrain was trained under a one-shot episodic protocol: each episode contains exactly one labeled positive support slice for the selected target region. During training and validation, support and query slices were drawn from different cases. Images were resized to \(256\times256\). The model was trained for 150 epochs, with 3,600 training episodes and 720 validation episodes per epoch and a mini-batch size of four. AdamW used an initial learning rate of \(10^{-4}\), weight decay \(10^{-4}\), and numerical-stability constant \(10^{-6}\). The learning rate was linearly warmed up for eight epochs and cosine-annealed to \(10^{-6}\). Gradients were clipped to a maximum norm of 0.5. Mixed precision and an exponential moving average of model parameters with decay 0.995 were used during optimization and evaluation.

The descriptor module used 100 primary FG descriptors and 300 primary BG descriptors. As defined in Section~\ref{sec:descriptor_adaptation}, two additional support-derived descriptors were appended to each FG and BG descriptor set. The descriptor grid size was 64, and the cosine-similarity scale was 20.

To improve robustness when a queried region is absent, TC and ET episodes included hard-negative and fully negative queries. A hard-negative query contains WT tissue but no voxel of the selected target region, whereas a fully negative query contains no WT tissue. The hard-negative and fully negative sampling ratios were \(0.20\) and \(0.10\), respectively, during training, and \(0.25\) and \(0.15\), respectively, during validation. Because a WT-empty slice contains no WT tissue by definition, WT has no distinct hard-negative category; its nominal hard-negative allocation is consequently sampled from the fully negative pool.

Model selection was based on the mean Dice score of reconstructed three-dimensional validation volumes. The selected checkpoint was then evaluated using the final volume-level protocol described below.

\subsection{Volume-Level Evaluation and Post-Processing}
\label{sec:evaluation_protocol}

Although FSS-UBrain performs two-dimensional support--query inference, all principal results were computed after reconstructing three-dimensional prediction volumes. Each query slice was processed independently, its predicted probability map was resized to the original in-plane resolution, and the maps were stacked along the original slice axis.

For held-out BraTS20 evaluation, support slices were drawn exclusively from the BraTS20 training partition. Five deterministic one-shot support draws were evaluated for each test volume.

For each retained external cohort, a fixed target-domain support pool of ten labeled cases was selected once using seed 1337 after overlap auditing. These cases were excluded from query evaluation, yielding fixed query cohorts of 85 cases for BraTS-Africa and 1,241 cases for BraTS23.

For BraTS-Africa, five deterministic one-shot support draws were constructed from the fixed support pool. For each draw and tumor region, one positive support slice was evaluated against the same 85-case query cohort. Region-level cohort means were first computed within each draw, after which the WT, TC, and ET results were averaged without weighting. Values are reported as mean \(\pm\) sample standard deviation across the five draws.

For BraTS23, two deterministic one-shot support draws, corresponding to seeds 1337 and 1338, were constructed from the fixed target-support pool. For each draw and tumor region, one positive support slice was evaluated against the same 1,241-case query cohort. Region-level cohort means were first computed within each draw, after which the WT, TC, and ET results were averaged without weighting. Values are reported as mean \(\pm\) sample standard deviation across the two draws.

The support pool, query cohort, and evaluated support configurations were shared across all compared methods within each experiment. No target-domain fine-tuning, gradient update, test-time parameter update, or query-label adaptation was performed.

All compared models used the same final inference protocol. Flip-based test-time augmentation averaged predictions from the original query slice, the horizontal flip, the vertical flip, and the simultaneous horizontal--vertical flip. The reconstructed probability volumes were smoothed only along the through-plane \(z\)-axis using a one-dimensional Gaussian filter with standard deviation \(\sigma_z=1.0\), measured in slices. Region-specific decision thresholds were selected exclusively on a fixed 12-case subset of the BraTS20 validation partition. Candidate thresholds were evaluated on the grid \(\{0.20,0.25,0.30,0.35,0.40,0.45,0.50,0.55,0.60,0.65,0.70\}\) under five deterministic support configurations (seeds 1337--1341), flip-based test-time augmentation, through-plane Gaussian smoothing, and nested-region post-processing. Two coordinate-wise passes were used to optimize WT, TC, and ET thresholds; the selected values were then fixed for held-out and external evaluation.

After region-specific thresholding, nested-region post-processing is applied to ensure anatomically consistent tumor predictions. The tumor core prediction is restricted to the predicted whole-tumor region, and the enhancing-tumor prediction is subsequently restricted to the resulting tumor-core region. Consequently, the final enhancing-tumor region is always contained within the tumor core, and the tumor core is always contained within the whole-tumor region. This procedure removes hierarchically inconsistent voxels while preserving the clinical nesting relationship among the three BraTS regions.

\subsection{Evaluation Metrics and Fair Comparison}
\label{sec:metrics_fairness}

\begin{table*}[!t]
\setlength{\abovedisplayskip}{0pt}
\setlength{\belowdisplayskip}{0pt}
\setlength{\abovedisplayshortskip}{0pt}
\setlength{\belowdisplayshortskip}{0pt}
\centering
\caption{Definitions, main advantages, and main limitations of the volume-level evaluation metrics.}
\label{tab:evaluation_metrics}
\renewcommand{\arraystretch}{1.2}
\setlength{\tabcolsep}{5pt}
% \resizebox{0.9\textwidth}{!}{%
% \scriptsize
\footnotesize
\begin{tabular}{|p{8.5cm}|p{4.5cm}|p{4cm}|}
\hline
\centering\textbf{Evaluation Metric and Formula}\arraybackslash
& \centering\textbf{Main Advantages}\arraybackslash
& \centering\textbf{Main Limitations}\arraybackslash\\
\hline
\hline
\textbf{Dice coefficient.} This overlap measure is defined as \cite{milletari2016v,taha2015metrics}
\begin{align}
    \operatorname{Dice}(M_r,\widehat{M}_r) = \frac{2TP+\epsilon_3}{2TP+FP+FN+\epsilon_3},
\end{align}
where \(\epsilon_3=10^{-6}\) is the small constant used for numerical stability. 
Dice is the voxel-wise \(F_1\) score of the predicted FG region and is reported as a percentage.
&
Primary overlap measure for imbalanced binary segmentation; weights \(FP\) and \(FN\) equally while excluding \(TN\), so it is not inflated by the large BG volume; comparable in scale with the BraTS literature.
& Invariant to error location; merges over- and under-segmentation into one value; unstable for sparse regions such as ET; degenerate for jointly empty pairs.
% &
% \centering\cite{milletari2016v,taha2015metrics}\arraybackslash 
\\
\hline
\textbf{HD95.} The 95th-percentile Hausdorff distance is defined as \cite{taha2015metrics}
\begin{align}
    \operatorname{HD95}(M_r,\widehat{M}_r)
=
Q_{0.95}\!\left(
\mathcal{D}(\partial M_r,\partial\widehat{M}_r)
\uplus
\mathcal{D}(\partial\widehat{M}_r,\partial M_r)
\right),
\end{align}
where \(\uplus\) denotes concatenation with multiplicities preserved, \(Q_{0.95}\) returns the 95th percentile of the resulting distance collection,
whereas \(\partial M_r\) and \(\partial\widehat{M}_r\) denote the surfaces of the reference and predicted masks, respectively. For two surfaces \(A\) and \(B\), \(\mathcal D(A,B)\) denotes the collection of directed nearest-surface distances from \(A\) to \(B\).
% The percentile is taken over the pooled bidirectional collection for all compared methods, and values are reported in millimeters.
&
Robust surface-distance measure in physical space that reduces sensitivity to isolated outlier voxels; interpretable as a clinical margin; sensitive to the geometric placement of errors that overlap metrics cannot express.
& Undefined when exactly one mask is empty, yielding a finite-pair summary over a varying subset; conveys no volumetric information; depends on voxel spacing; the discarded \(5\%\) tail may hide extreme deviations.
% &
% \centering\cite{taha2015metrics}\arraybackslash 
\\
\hline
\textbf{Intersection over union (IoU).} The overlap between prediction and reference is \cite{taha2015metrics}
\begin{align}
    \operatorname{IoU}(M_r,\widehat{M}_r)
=
\frac{TP}{TP+FP+FN}.
\end{align}
Also known as the Jaccard index, it satisfies \(\operatorname{IoU}=\operatorname{Dice}/(2-\operatorname{Dice})\) for an individual volume--region pair, with both quantities expressed as fractions.
&
Supplementary overlap measure that penalizes false positives and false negatives symmetrically; stricter than Dice, hence separates methods whose Dice values are compressed near the upper end of the scale.
& Monotone transform of Dice per pair and therefore largely redundant; shares the insensitivity of Dice to error location; absolute values are not comparable with Dice-based results reported elsewhere.
% &
% \centering\cite{taha2015metrics}\arraybackslash 
\\
\hline
\textbf{Precision.} The reliability of predicted FG voxels is given by \cite{taha2015metrics}
\begin{align}
    \operatorname{Precision}(M_r,\widehat{M}_r)
=
\frac{TP}{TP+FP}.
\end{align}
It is equivalently the positive predictive value.
&
Proportion of predicted FG voxels that are correct; isolates the \(FP\) component that Dice merges with \(FN\), thereby quantifying over-segmentation.
& Driven close to its maximum by conservative predictions, hence uninformative unless read jointly with recall; degenerate on target-absent pairs; disregards the spatial distribution of false positives.
% &
% \centering\cite{taha2015metrics}\arraybackslash 
\\
\hline

\textbf{Recall.} The recovery of reference FG voxels is \cite{taha2015metrics}
\begin{align}
    \operatorname{Recall}(M_r,\widehat{M}_r)
=
\frac{TP}{TP+FN}.
\end{align}
It is equivalently the sensitivity or true-positive rate.
&
Proportion of reference FG voxels recovered by the prediction; quantifies under-segmentation, the error type with the greatest clinical cost in surgical and radiotherapy planning.
& Attains its maximum for any model labeling the entire brain as FG; entirely insensitive to false positives; degenerate on target-absent pairs.
% &
% \centering\cite{taha2015metrics}\arraybackslash 
\\
\hline
\textbf{Specificity.} The correct rejection of BG voxels is \cite{taha2015metrics}
\begin{align}
    \operatorname{Specificity}(M_r,\widehat{M}_r)
=
\frac{TN}{TN+FP}.
\end{align}
It is equivalently the true-negative rate.
&
Proportion of BG voxels correctly rejected as BG; describes the negative class, which the overlap metrics exclude; provides a volume-wide check against indiscriminate FG activation.
& Saturates near \(100\%\) because BG voxels dominate the volume, hence offers little resolution among models; a high value may still correspond to a clinically significant number of false-positive voxels.
% &
% \centering\cite{taha2015metrics}\arraybackslash 
\\
\hline
\textbf{FP-neg-rate.} For a volume--region pair with an empty reference mask, FP-neg-rate is defined as the fraction of all voxels that are incorrectly predicted as FG. Mathematically, FP-neg-rate is expressed as \cite{hartmann2023mism}
\begin{align}
    \operatorname{FP\text{-}neg\text{-}rate}(M_r,\widehat{M}_r)
=
\frac{FP}{FP+TN},
\qquad
\lvert M_r\rvert=0,
\end{align}
where \(\lvert M_r\rvert\) denotes the number of FG voxels in the reference volume mask.
&
False-positive FG fraction on volume--region pairs whose reference mask is empty; the only metric in the panel that isolates target-absent reliability, on which the overlap metrics are degenerate; directly probes the failure mode that negative-query training is designed to suppress.
& Defined only on the empty-reference subset, so regional averages rest on unequal sample sizes and are dominated by TC and ET; blind to the spatial clustering and clinical severity of the false positives; attains a perfect value for a model that predicts no FG voxel anywhere.
% &
% \centering \cite{hartmann2023mism}\arraybackslash 
\\
\hline
\end{tabular}%
% }
\end{table*}

All metrics in \autoref{tab:evaluation_metrics} were computed from final reconstructed and post-processed three-dimensional masks. For each clinical region \(r\in\{\mathrm{WT},\mathrm{TC},\mathrm{ET}\}\), \(M_r\) denotes the binary reference volume mask and \(\widehat{M}_r\) denotes the corresponding final predicted binary volume mask; the hat indicates a model prediction. The voxel-wise counts \(TP\), \(FP\), \(TN\), and \(FN\) denote true positives, false positives, true negatives, and false negatives, respectively.
% For Dice, \(\epsilon=10^{-6}\) is used for numerical stability.

Each of these metrics characterizes segmentation quality only with respect to a specific property, and each is therefore blind to, or potentially misleading about, some other property. Overlap metrics quantify how much of the tumor volume is recovered but are invariant to the spatial arrangement of the errors; surface-distance metrics describe boundary geometry but convey no information about the amount of correctly labeled volume and are undefined when exactly one mask is empty; precision and recall isolate the two error types that Dice aggregates, yet either one can be improved at the direct expense of the other; and background-oriented metrics saturate because background voxels dominate a brain MRI volume. Reporting a single figure of merit would consequently allow a model to appear strong while failing in a manner that the selected metric cannot express. This is the reason for reporting a panel of metrics rather than a single score. Accordingly, \autoref{tab:evaluation_metrics} states, for every metric, both the property that it captures and the property that it cannot capture. The panel is assembled so that no limitation of one metric is shared by all of the others: Dice and IoU quantify volumetric agreement, HD95 adds boundary geometry in physical units, precision and recall decompose the error into over- and under-segmentation, specificity summarizes background rejection, and FP-neg-rate isolates the target-absent behavior that none of the preceding metrics can represent. The limitations listed in the last column of \autoref{tab:evaluation_metrics} are consequently covered pairwise: the insensitivity of Dice and IoU to error location is covered by HD95; the exclusion of empty-reference pairs from HD95 is covered by FP-neg-rate; the conservative-prediction bias of precision is covered by recall and the over-segmentation bias of recall by precision, with Dice summarizing their balance; and the saturation of specificity under BG dominance is covered by FP-neg-rate, which reports the complementary false-positive fraction restricted to the pairs whose reference mask is empty. Dice and HD95 are reported for every method in the comparative experiments, whereas the remaining metrics are reported for the negative-query sensitivity analysis of Section~\ref{sec:negative_ratio_analysis}, in which the trade-off between foreground recovery and target-absent false-positive suppression is examined explicitly.

% For HD95,   FP-neg-rate is evaluated only for volume--region pairs with an empty reference mask; on such a pair, it is the fraction of all voxels that are incorrectly predicted as FG.
% Requires: \usepackage{amsmath,graphicx,array}

For jointly empty prediction--reference pairs, Dice, IoU, precision, and recall were assigned a value of one, whereas HD95 was assigned a value of zero. When the reference mask was non-empty but the prediction was empty, precision was assigned a value of zero. When the reference mask was empty but the prediction was non-empty, recall was assigned a value of zero. When exactly one mask was empty, HD95 was undefined and excluded from the finite-pair HD95 summary. Consequently, the reported HD95 values are conditional on this empty-mask convention and should be interpreted jointly with the corresponding Dice results. Specificity was assigned a value of one when its denominator was zero. These conventions were applied identically to all evaluated methods and configurations. Because these conventions determine the reported values in exactly the degenerate cases identified in the third column of \autoref{tab:evaluation_metrics}, the metrics should be interpreted as a coordinated panel rather than individually, and the comparative claims in Section~\ref{Sect:5} are reported together with the complementary metrics of the panel wherever these are available.

For the extended metric analysis, IoU, precision, recall, and specificity were computed separately for each reconstructed test volume, tumor region, and one-shot support draw. For each region, the reported value was obtained by averaging over all test-volume--support-draw pairs. The corresponding mean value was subsequently calculated as the unweighted mean of the WT, TC, and ET regional results.

FP-neg-rate was evaluated only on volume--region--support-draw pairs with an empty reference mask. Within each eligible region, false-positive and total voxel counts were pooled before computing the regional rate. The reported mean rate was then obtained as the unweighted average over eligible regions; WT contributed only when a WT-empty reference volume was present.

Fairness was enforced through a common BraTS20 split, region-mask construction, preprocessing pipeline, input resolution, validation-only threshold calibration, volume reconstruction, post-processing procedure, and metric implementation. Within each experiment, support-conditioned methods used the same fixed support pool, one-shot support configurations, and query cohort. Methods that do not require a support image were evaluated through the same query-volume pipeline without a support input. The common query-volume pipeline controls data handling and evaluation. The labeled support input remains an explicit requirement of support-conditioned methods and is not used by non-FSS models; this model-specific input requirement is reported as part of the comparison rather than treated as a shared input.

\section{Numerical Results}
\label{Sect:5}

Unless otherwise stated, all results were obtained from reconstructed three-dimensional volumes using the protocol in Section~\ref{sec:evaluation_protocol}. All dimensionless metrics, including Dice, IoU, precision, recall, specificity, and FP-neg-rate, are reported as percentages, whereas HD95 is reported in millimeters. The primary FSS-UBrain configuration uses hard-/fully-negative sampling ratios of \(20/10\) during training and \(25/15\) during validation.

\subsection{Training and Validation Dynamics on BraTS20}
\label{sec:training_dynamics}

\begin{figure*}[t!]
\centering
\includegraphics[width=\textwidth]{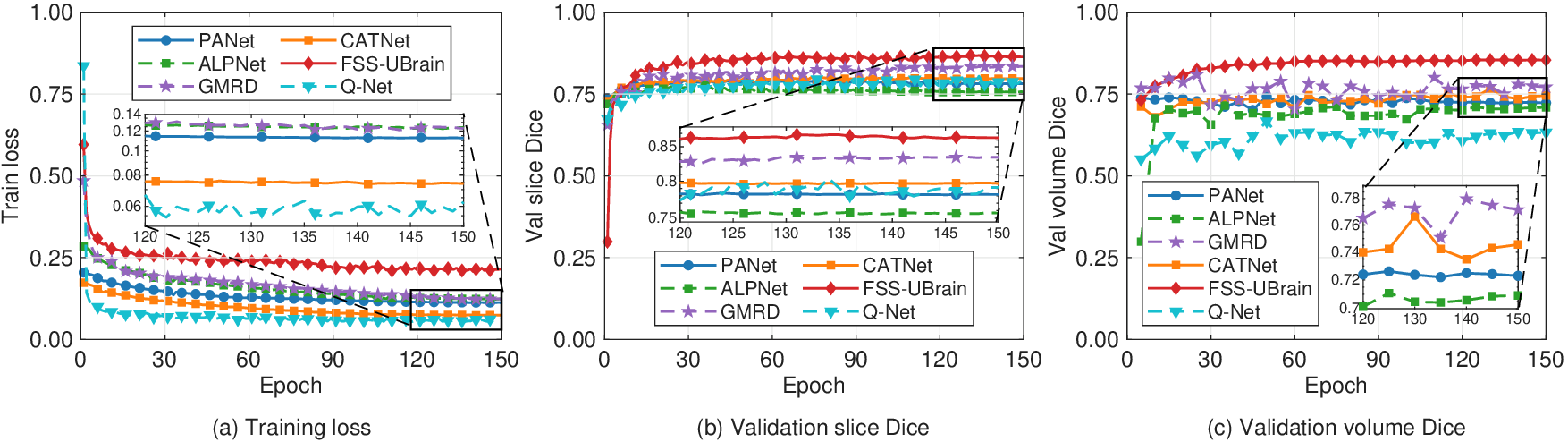}
\caption{Training and validation dynamics of the compared models on BraTS20.}
\label{Fig:evaluate_metrics}
\end{figure*}

\autoref{Fig:evaluate_metrics} summarizes training loss, validation slice Dice, and validation volume Dice. The losses decrease and the validation curves approach stable regimes under the selected schedules. Because the methods optimize different objectives, absolute loss values are not directly comparable across models. The volume-level validation curves identify the checkpoints used in the subsequent evaluations and are interpreted as optimization diagnostics rather than as evidence of comparative convergence. Checkpoint selection was based on reconstructed validation volumes rather than on isolated slice predictions.

\subsection{In-Domain Comparison with Few-Shot Baselines}
\label{sec:brats20_fss_benchmark}

\autoref{tab:BraTS20_results_fss} compares FSS-UBrain with ALPNet \cite{ouyang2020self}, PANet \cite{wang2019panet}, Q-Net \cite{shen2023q}, CATNet \cite{lin2023few}, and GMRD \cite{cheng2024few} on the held-out BraTS20 cohort. All methods use the shared volume-level protocol in Section~\ref{sec:evaluation_protocol}; for FSS-UBrain, the reported values correspond to the fixed primary hard-/fully-negative configuration of \(20/10\) during training and \(25/15\) during validation.

\begin{table*}[t]
\centering
\caption{BraTS20 few-shot results reported as mean $\pm$ sample standard deviation over five deterministic support draws. Best point estimates
are bold.}
\label{tab:BraTS20_results_fss}
\footnotesize
\renewcommand{\arraystretch}{1.18}
\setlength{\tabcolsep}{7pt}

\begin{tabular}{@{}l cc cc cc@{}}
\toprule
& \multicolumn{2}{c}{\textbf{WT}}
& \multicolumn{2}{c}{\textbf{TC}}
& \multicolumn{2}{c}{\textbf{ET}} \\
\cmidrule(lr){2-3}
\cmidrule(lr){4-5}
\cmidrule(lr){6-7}

\textbf{Model}
& \textbf{Dice}
& \textbf{HD95}
& \textbf{Dice}
& \textbf{HD95}
& \textbf{Dice}
& \textbf{HD95} \\
\midrule

ALPNet~\cite{ouyang2020self}
& $76.38 \pm 6.39$
& $41.48 \pm 3.86$
& $56.74 \pm 10.96$
& $47.80 \pm 18.76$
& $47.88 \pm 4.35$
& $36.36 \pm 23.24$ \\

PANet~\cite{wang2019panet}
& $80.44 \pm 5.38$
& $32.48 \pm 2.46$
& $75.33 \pm 2.06$
& $26.14 \pm 7.86$
& $62.65 \pm 2.78$
& $15.65 \pm 2.64$ \\

Q-Net~\cite{shen2023q}
& $82.74 \pm 0.58$
& $37.23 \pm 1.53$
& $66.53 \pm 10.30$
& $45.36 \pm 4.22$
& $55.06 \pm 3.23$
& $40.43 \pm 3.15$ \\

CATNet~\cite{lin2023few}
& $82.48 \pm 0.10$
& $37.70 \pm 0.98$
& $68.89 \pm 2.60$
& $39.01 \pm 14.16$
& $69.34 \pm 1.17$
& $18.08 \pm 5.25$ \\

GMRD~\cite{cheng2024few}
& $85.98 \pm 1.05$
& $26.13 \pm 3.64$
& $79.09 \pm 1.20$
& $24.16 \pm 5.95$
& $68.95 \pm 7.75$
& $15.57 \pm 9.13$ \\

FSS-UBrain (ours)
& $\mathbf{89.82 \pm 0.03}$
& $\mathbf{11.12 \pm 0.08}$
& $\mathbf{82.14 \pm 4.35}$
& $\mathbf{9.01 \pm 0.70}$
& $\mathbf{77.42 \pm 3.42}$
& $\mathbf{4.46 \pm 1.63}$ \\
\bottomrule
\end{tabular}
\end{table*}
FSS-UBrain achieves the highest Dice and lowest HD95 point estimates for all three regions in \autoref{tab:BraTS20_results_fss}. Its Dice scores are \(89.82 \pm 0.03\), \(82.14 \pm 4.35\), and \(77.42 \pm 3.42\) for WT, TC, and ET, respectively, with corresponding HD95 values of \(11.12 \pm 0.08\), \(9.01 \pm 0.70\), and \(4.46 \pm 1.63\)~mm. The largest Dice point-estimate margin occurs for ET, where FSS-UBrain exceeds the strongest competing few-shot baseline by \(8.08\) percentage points. The lower HD95 point estimates indicate that the overlap improvements are accompanied by closer boundary agreement.

\subsection{BraTS20 Comparison with Non-FSS Baselines}
\label{sec:brats20_nonfss_benchmark}

\autoref{tab:BraTS20_results_non_fss} compares the same fixed FSS-UBrain configuration with U-Net \cite{ronneberger2015unet}, UNet++ \cite{zhou2019unet++}, TransUNet \cite{chen2021transunet}, SwinUNet \cite{cao2022swin}, VMUNet \cite{ruan2024vm}, and the Segment Anything Model (SAM) \cite{kirillov2023segment} on the held-out BraTS20 query cohort. All reported comparisons use the cohort, support, reconstruction, and metric definitions specified in Section~\ref{sec:evaluation_protocol}.

\begin{table}[t]
\centering
\caption{BraTS20 volume-level Dice and HD95 for non-FSS baselines and FSS-UBrain. The best point estimate in each column is shown in bold.}
\label{tab:BraTS20_results_non_fss}
\scriptsize
\renewcommand{\arraystretch}{1.15}
\setlength{\tabcolsep}{3pt}

\resizebox{\linewidth}{!}{%
\begin{tabular}{@{}l cc cc cc@{}}
\toprule
& \multicolumn{2}{c}{\textbf{WT}}
& \multicolumn{2}{c}{\textbf{TC}}
& \multicolumn{2}{c}{\textbf{ET}} \\
\cmidrule(lr){2-3}
\cmidrule(lr){4-5}
\cmidrule(lr){6-7}

\textbf{Model}
& \textbf{Dice}
& \textbf{HD95}
& \textbf{Dice}
& \textbf{HD95}
& \textbf{Dice}
& \textbf{HD95} \\
\midrule

U-Net~\cite{ronneberger2015unet}
& 86.20 & 28.01
& 80.20 & 25.78
& 75.72 & 17.65 \\

UNet++~\cite{zhou2019unet++}
& 86.35 & 27.61
& 80.61 & 27.68
& 74.77 & 23.74 \\

TransUNet~\cite{chen2021transunet}
& 82.72 & 40.91
& 75.24 & 39.16
& 70.43 & 16.79 \\

SwinUNet~\cite{cao2022swin}
& 84.29 & 24.24
& 74.01 & 23.28
& 68.23 & 19.07 \\

VMUNet~\cite{ruan2024vm}
& 88.32 & 18.41
& 80.77 & 20.97
& 75.09 & 13.44 \\

SAM~\cite{kirillov2023segment}
& 86.06 & \textbf{4.25}
& \textbf{84.83} & \textbf{4.86}
& 68.74 & 4.75 \\

FSS-UBrain (ours)
& \textbf{89.82} & 11.12
& 82.14 & 9.01
& \textbf{77.42} & \textbf{4.46} \\
\bottomrule
\end{tabular}%
}
\end{table}

FSS-UBrain achieves the highest WT Dice (\(89.82\)) and ET Dice (\(77.42\)) in \autoref{tab:BraTS20_results_non_fss}. It also attains the lowest ET HD95 (\(4.46\)~mm). SAM~\cite{kirillov2023segment} achieves the strongest TC Dice (\(84.83\)) and the lowest WT and TC HD95 (\(4.25\) and \(4.86\)~mm), respectively. Therefore, the comparison does not indicate uniform dominance by a single model across all regions and metrics. Rather, FSS-UBrain provides the strongest WT and ET overlap and the most accurate ET boundary distance among the listed methods, while SAM remains stronger on TC and on the WT/TC surface-distance measures.

\subsection{Target-Supported Cross-Cohort Evaluation}
\label{sec:cross_cohort_evaluation}

\autoref{tab:cross_dataset_results} evaluates transfer beyond the BraTS20 source cohort. For each external cohort, ten labeled target cases were reserved as a case-disjoint support pool and excluded from query evaluation. BraTS-Africa used five deterministic support draws on the remaining 85 cases, whereas BraTS23 used two deterministic support draws on the remaining 1,241 cases.

\begin{table}[t]
\centering
\caption{Target-supported cross-cohort mean results. BraTS-Africa values are mean $\pm$ sample standard deviation over five deterministic support draws; BraTS23 values are mean $\pm$ sample standard deviation over two deterministic support draws (seeds 1337 and 1338). Best point estimates are bold.}
\label{tab:cross_dataset_results}
\footnotesize
\renewcommand{\arraystretch}{1.15}
\setlength{\tabcolsep}{3pt}

\resizebox{\linewidth}{!}{%
\begin{tabular}{@{}l cc cc@{}}
\toprule
& \multicolumn{2}{c}{\textbf{BraTS-Africa}}
& \multicolumn{2}{c}{\textbf{BraTS23}} \\
\cmidrule(lr){2-3}\cmidrule(lr){4-5}
\textbf{Model}
& \textbf{Mean Dice}
& \textbf{Mean HD95}
& \textbf{Mean Dice}
& \textbf{Mean HD95} \\
\midrule

ALPNet~\cite{ouyang2020self}
& $66.83 \pm 0.85$
& $40.40 \pm 2.05$
& $69.15 \pm 4.32$
& $40.50 \pm 4.10$ \\

PANet~\cite{wang2019panet}
& $74.37 \pm 1.08$
& $16.29 \pm 1.01$
& $76.77 \pm 0.28$
& $21.94 \pm 1.17$ \\

Q-Net~\cite{shen2023q}
& $68.83 \pm 6.04$
& $28.94 \pm 0.43$
& $71.02 \pm 0.76$
& $42.39 \pm 0.02$ \\

CATNet~\cite{lin2023few}
& $70.65 \pm 2.01$
& $26.56 \pm 1.95$
& $71.34 \pm 0.43$
& $36.21 \pm 2.61$ \\

GMRD~\cite{cheng2024few}
& $77.65 \pm 0.86$
& $18.46 \pm 1.14$
& $81.84 \pm 0.98$
& $13.44 \pm 1.01$ \\

FSS-UBrain (ours)
& $\mathbf{78.04 \pm 0.77}$
& $\mathbf{11.03 \pm 0.43}$
& $\mathbf{83.41 \pm 0.26}$
& $\mathbf{7.86 \pm 0.05}$ \\

\bottomrule
\end{tabular}%
}
\end{table}

FSS-UBrain achieves the highest mean Dice and lowest finite-pair mean HD95 point estimates on both external cohorts. On BraTS-Africa, it obtains \(78.04\pm0.77\) Dice and \(11.03\pm0.43\)~mm HD95 across five support draws. On BraTS23, it obtains \(83.41\pm0.26\) Dice and \(7.86\pm0.05\)~mm HD95 across two support draws.

\subsection{Effect of Source Versus Target Support on BraTS-Africa}
\label{sec:africa_support_source}

\autoref{tab:brats_africa_support_mean_results} examines how the provenance of the labeled support examples affects cross-cohort inference. Source support refers to support slices drawn from the BraTS20 training partition, whereas target support refers to support slices drawn from reserved, case-disjoint target-domain cases. In this BraTS-Africa analysis, target support is drawn from a reserved pool of ten labeled BraTS-Africa cases. In both support conditions, the ten reserved BraTS-Africa cases are excluded from the query set, leaving the same 85-case BraTS-Africa query cohort for every method and support condition. Five deterministic one-shot support draws are evaluated under each condition. Thus, the comparison changes only the provenance of support evidence while preserving the query cases, inference pipeline, post-processing, and metric aggregation.

\begin{table}[t]
\centering
\caption{BraTS-Africa mean results under source and target support (mean $\pm$ sample standard deviation over five draws; same 85 query cases). Best point estimates within each support condition are bold.}
\label{tab:brats_africa_support_mean_results}
\scriptsize
\renewcommand{\arraystretch}{1.15}
\setlength{\tabcolsep}{2pt}

\resizebox{\linewidth}{!}{%
\begin{tabular}{@{}l cc cc@{}}
\toprule
& \multicolumn{2}{c}{\textbf{Source support}}
& \multicolumn{2}{c}{\textbf{Target support}} \\
\cmidrule(lr){2-3}
\cmidrule(lr){4-5}

\textbf{Model}
& \textbf{Mean Dice}
& \textbf{Mean HD95}
& \textbf{Mean Dice}
& \textbf{Mean HD95} \\
\midrule

ALPNet~\cite{ouyang2020self}
& $61.32 \pm 2.82$
& $38.07 \pm 11.75$
& $66.83 \pm 0.85$
& $40.40 \pm 2.05$ \\

PANet~\cite{wang2019panet}
& $73.05 \pm 1.71$
& $16.90 \pm 0.77$
& $74.37 \pm 1.08$
& $16.29 \pm 1.01$ \\

Q-Net~\cite{shen2023q}
& $70.60 \pm 2.49$
& $29.45 \pm 1.29$
& $68.83 \pm 6.04$
& $28.94 \pm 0.43$ \\

CATNet~\cite{lin2023few}
& $72.59 \pm 1.21$
& $25.82 \pm 3.46$
& $70.65 \pm 2.01$
& $26.56 \pm 1.95$ \\

GMRD~\cite{cheng2024few}
& $\mathbf{76.62 \pm 2.18}$
& $18.33 \pm 2.43$
& $77.65 \pm 0.86$
& $18.46 \pm 1.14$ \\

FSS-UBrain (ours)
& $74.55 \pm 2.47$
& $\mathbf{13.03 \pm 1.33}$
& $\mathbf{78.04 \pm 0.77}$
& $\mathbf{11.03 \pm 0.43}$ \\
\bottomrule
\end{tabular}%
}
\end{table}

For FSS-UBrain, replacing source support with target support increases the mean Dice point estimate from \(74.55 \pm 2.47\) to \(78.04 \pm 0.77\) and reduces the mean HD95 point estimate from \(13.03 \pm 1.33\) to \(11.03 \pm 0.43\)~mm. Under target support, FSS-UBrain achieves the highest mean Dice point estimate and the lowest mean HD95 point estimate among the compared methods. Because the query cohort, model parameters, and inference procedure are unchanged, these differences are associated with support provenance under the stated protocol.

The effect of target support is method dependent. Target support increases the mean Dice point estimate for ALPNet~\cite{ouyang2020self}, PANet~\cite{wang2019panet}, GMRD~\cite{cheng2024few}, and FSS-UBrain but decreases it for Q-Net~\cite{shen2023q} and CATNet~\cite{lin2023few}. HD95 changes are likewise non-uniform. These results indicate that support provenance can alter one-shot predictions, but they do not establish universal superiority of target support.

\subsection{Component Ablation on BraTS20}
\label{sec:component_ablation}

\autoref{tab:ablation_brats20} quantifies the contribution of the main components on BraTS20. DMP denotes the descriptor-matching pathway of the support-conditioned descriptor-adaptation block described in Section~\ref{sec:descriptor_adaptation}, which performs multi-descriptor FG--BG matching; SGB denotes the support-guided bottleneck, PC denotes the pyramid-context module, and BR denotes boundary-guided refinement.

\begin{table}[t!]
\captionsetup{justification=raggedright,singlelinecheck=false}
\caption{BraTS20 mean component-ablation results across WT, TC, and ET.
Best point estimates are bold.}
\label{tab:ablation_brats20}
\centering
\scriptsize
\renewcommand{\arraystretch}{1.12}
\setlength{\tabcolsep}{2.2pt}

\resizebox{0.8\columnwidth}{!}{%
\begin{tabular}{@{}l c c c c c@{}}
\toprule
\textbf{Metric}
& \textbf{Full model}
& \textbf{w/o DMP}
& \textbf{w/o SGB}
& \textbf{w/o PC}
& \textbf{w/o BR} \\
\midrule

Mean Dice
& \textbf{83.13}
& 82.33
& 82.42
& 82.75
& 82.98 \\

Mean HD95
& \textbf{8.20}
& 9.27
& 9.96
& 9.86
& 9.59 \\
\bottomrule
\end{tabular}%
}
\end{table}

The full model obtains the highest mean Dice (\(83.13\)) and the lowest mean HD95 (\(8.20\)~mm). Removing DMP yields the largest observed Dice reduction (\(0.80\) percentage points), supporting the contribution of multi-descriptor matching under the evaluated configuration. Removing SGB produces the largest HD95 increase (\(1.76\)~mm), while the removal of PC or BR also degrades both overlap and boundary accuracy. These point-estimate changes are consistent with complementary contributions from descriptor matching, support interaction, contextual aggregation, and boundary refinement under the evaluated configuration.

\subsection{Qualitative Comparison}
\label{sec:qualitative_comparison}

\autoref{tab:qualitative_comparison} provides illustrative support--query examples for WT, TC, and ET across the compared few-shot methods. Each block shows the support FLAIR image and its corresponding mask, followed by the query FLAIR image, query reference mask, and prediction. The examples complement the volume-level results by illustrating the spatial behavior of the competing methods.

\newcommand{\qualblockheader}{%
\scriptsize
\begin{tabular*}{\linewidth}{@{\extracolsep{\fill}}ccccc@{}}
\textbf{SP FLAIR} &
\textbf{SP Mask} &
\textbf{Query FLAIR} &
\textbf{Query GT} &
\textbf{Query Pred}
\end{tabular*}
}

\newcommand{\rowlabels}{%
\scriptsize
\begin{tabular}{@{}c@{}}
\textbf{WT}\\[1.25cm]
\textbf{TC}\\[1.25cm]
\textbf{ET}
\end{tabular}
}

\newcommand{\qualblock}[2]{%
\begin{minipage}[t]{0.42\textwidth}
\centering
\textbf{#1}\\[1mm] \qualblockheader\\[1mm] \fbox{\includegraphics[width=\linewidth]{#2}}
\end{minipage}
}

\newcommand{\qualpair}[4]{%
\begin{minipage}[t]{0.035\textwidth}
\centering
\vspace{1.1cm}
\rowlabels
\end{minipage}
& \qualblock{#1}{#2} & \qualblock{#3}{#4} }

\begin{figure*}[t!]
\centering
\setlength{\tabcolsep}{4pt}
\renewcommand{\arraystretch}{1.05}
\setlength{\fboxsep}{0pt}
\setlength{\fboxrule}{0.8pt}

\resizebox{0.90\textwidth}{!}{%
\begin{tabular}{@{}c c c@{}}
\toprule

\qualpair
{FSS-UBrain}{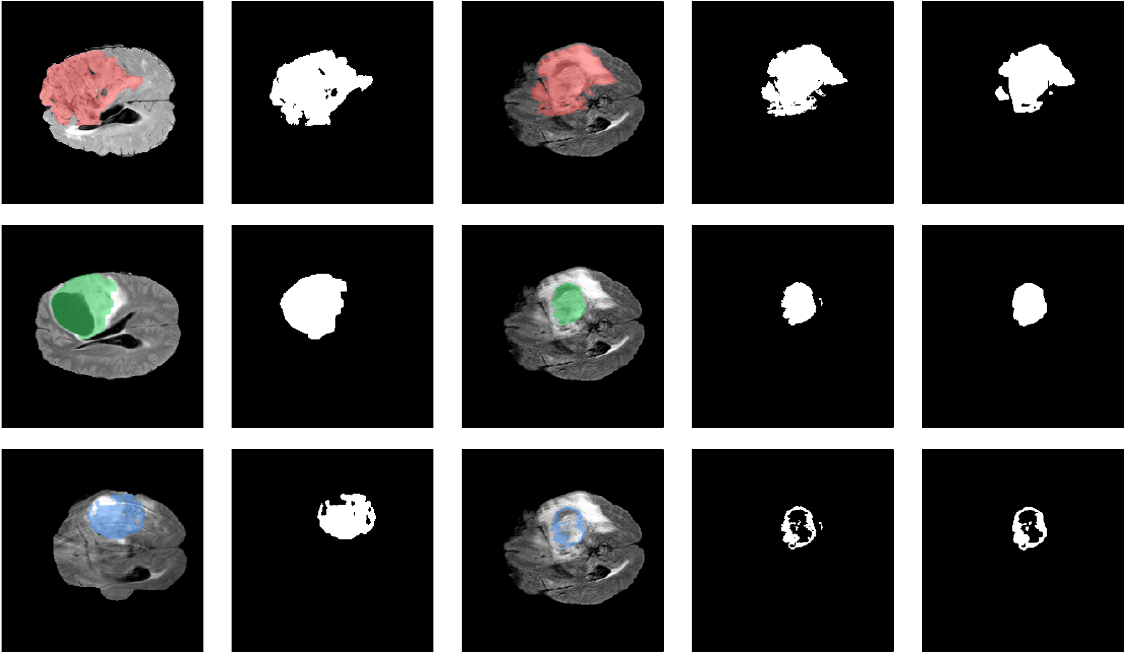}
{ALPNet}{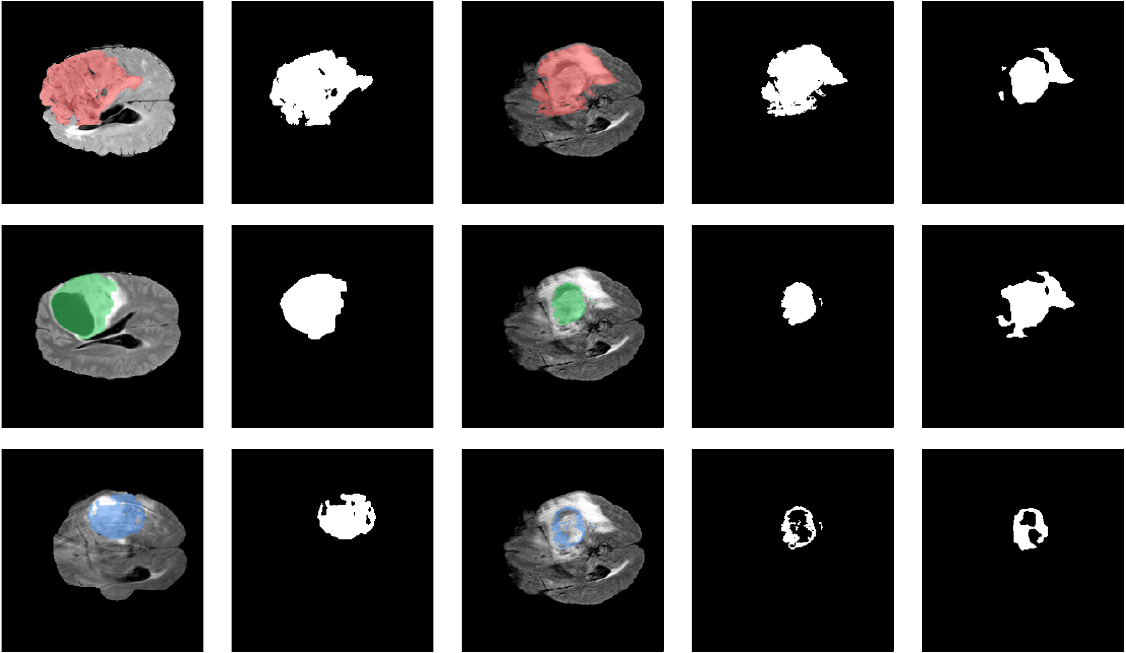}
\\[3mm]

\qualpair
{PANet}{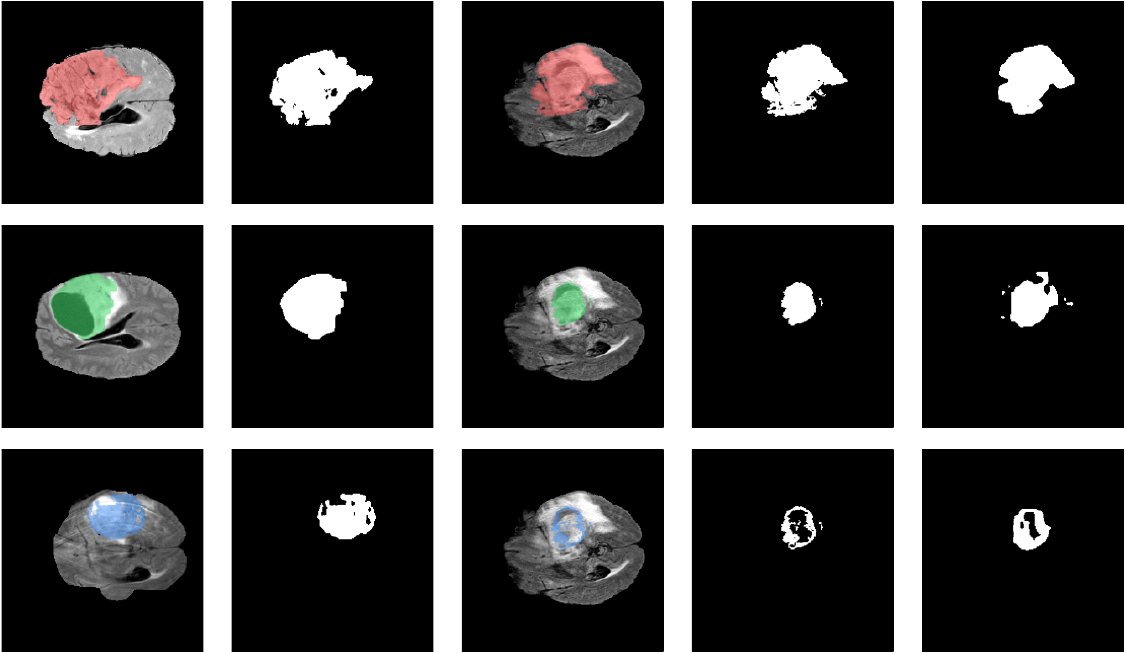}
{CATNet}{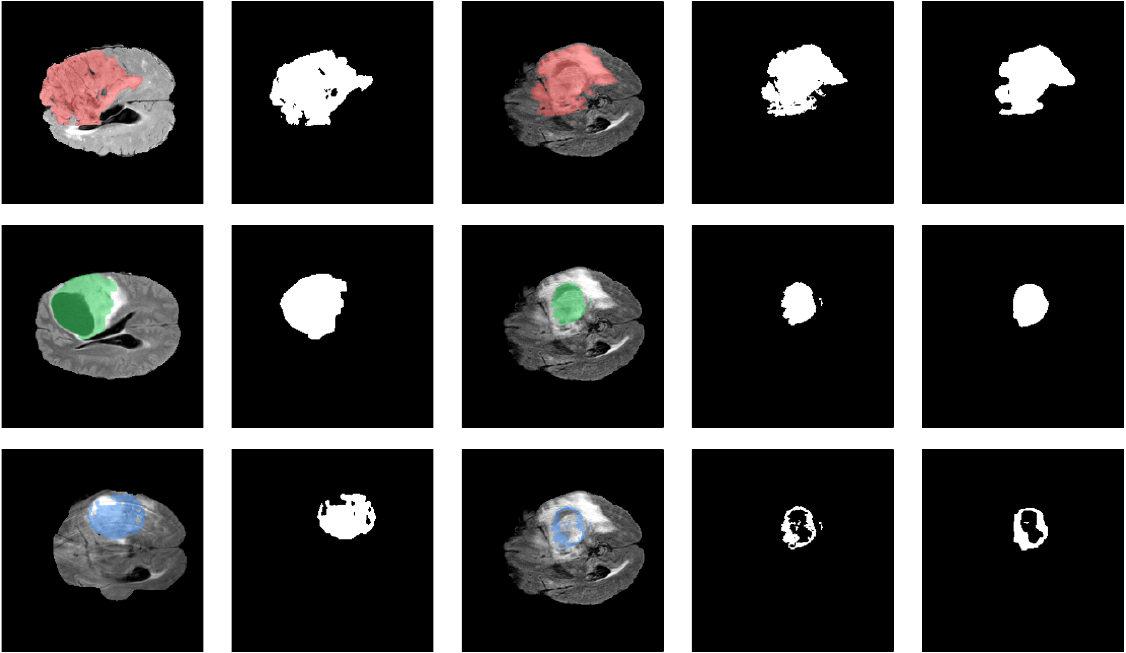}
\\[3mm]

\qualpair
{GMRD}{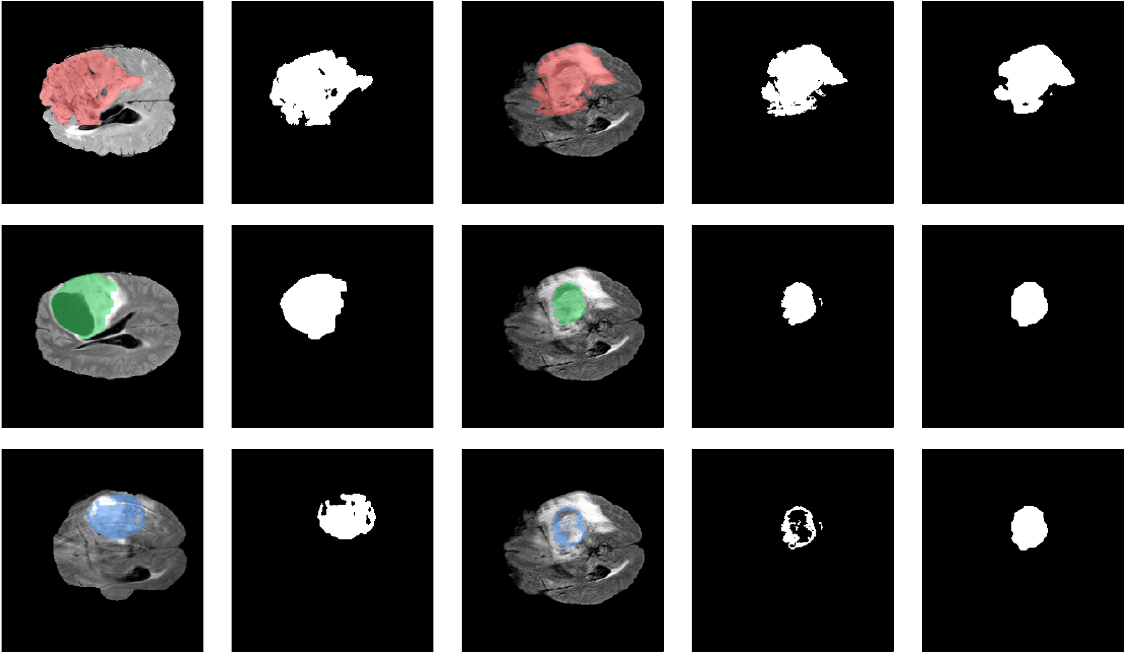}
{Q-Net}{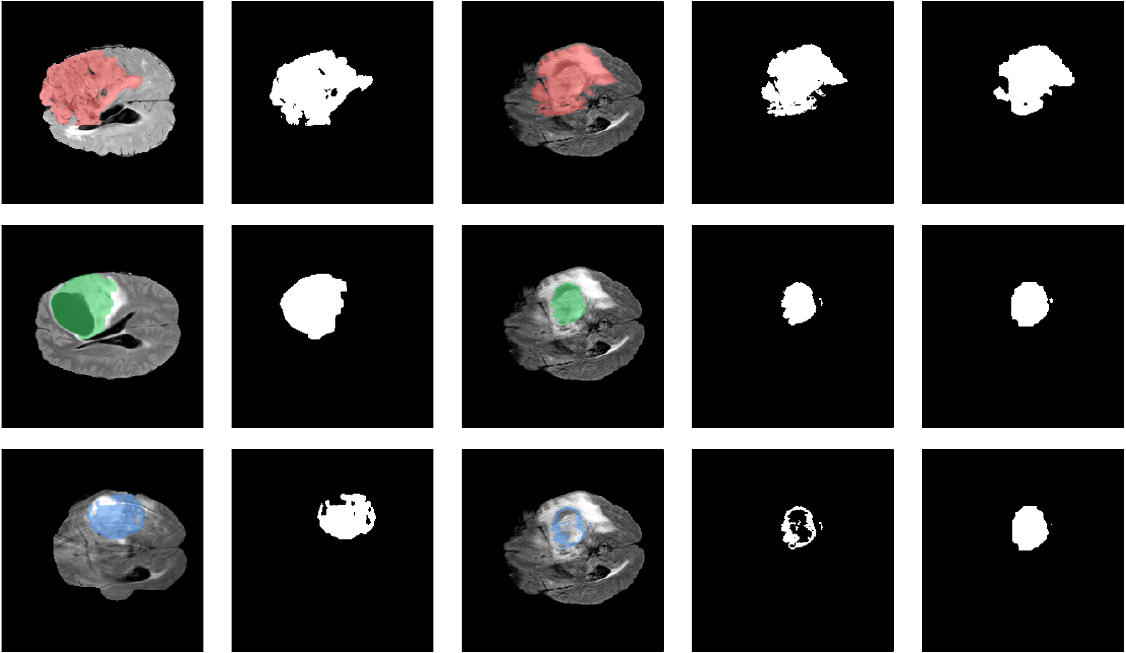}
\\

\bottomrule
\end{tabular}%
}
\caption{Qualitative few-shot comparisons using four-modality MRI input; only FLAIR is displayed. SP denotes support, GT ground truth, and Pred prediction.}
\label{tab:qualitative_comparison}
\end{figure*}

In the displayed examples, FSS-UBrain yields predictions that more closely follow the reference masks for the selected TC and ET cases, whereas some baseline predictions show incomplete recovery or displaced boundaries.

\subsection{Post-Hoc Sensitivity Analysis of Negative-Query Sampling}
\label{sec:negative_ratio_analysis}

\autoref{tab:pos_only_neg_ratio} reports a post-hoc sensitivity analysis of five negative-query training configurations, together with the fixed primary configuration used in the principal comparisons. All configurations were evaluated on the same held-out BraTS20 test split using five deterministic one-shot support draws. The analysis examines the empirical trade-off among volumetric overlap, boundary agreement, foreground sensitivity, background rejection, and false-positive activation on target-absent volumes. It does not replace the fixed primary \(20/10/25/15\) configuration used for the benchmark comparisons. The ratio columns are ordered as training hard-negative, training fully negative, validation hard-negative, and validation fully negative.

\begin{table*}[t!]
\centering
\footnotesize
\captionsetup{justification=raggedright,singlelinecheck=false}
\caption{BraTS20 sensitivity to hard- and fully negative-query sampling. Mean Dice and mean HD95 are unweighted averages across WT, TC, and ET; ratios denote training hard/full and validation hard/full sampling. Best point estimates are bold.}
\label{tab:pos_only_neg_ratio}
\renewcommand{\arraystretch}{1.15}

\resizebox{\linewidth}{!}{%
\begin{tabular}{@{}l c c c c c c c c c c c@{}}
\toprule
\textbf{Cohort}
& \textbf{Train Hard Neg}
& \textbf{Train Full Neg}
& \textbf{Val Hard Neg}
& \textbf{Val Full Neg}
& \textbf{Mean Dice}
& \textbf{Mean HD95}
& \textbf{Mean IoU}
& \textbf{Precision}
& \textbf{Recall}
& \textbf{Specificity}
& \textbf{FP-neg-rate} \\
\midrule

\multirow{6}{*}{BraTS20}
& 0  & 0  & 0  & 0  & 75.46 & 41.97 & 63.93 & 72.82 & 83.33 & 99.89 & 0.35  \\
& 20 & 0  & 25 & 0  & 82.72 & 10.68 & 73.91 & 82.11 & 86.99  & 99.93 & 0.16 \\
& 0  & 10 & 0  & 15 & 81.52 & 12.81 & 72.41 & 80.68  & 86.57 & 99.92 & 0.23 \\
& 10 & 5  & 15 & 10 & 83.21 & 11.42 & 74.27 & 82.09 & 87.06 & 99.93 & 0.15  \\
& 20 & 10 & 25 & 15
& 83.13
& 8.20
& 74.34
& 82.59
& 86.81
& 99.93
& \textbf{0.10} \\
& 30 & 15 & 35 & 20
& \textbf{84.19}
& \textbf{7.61}
& \textbf{75.90}
& \textbf{83.75}
& \textbf{87.45}
& \textbf{99.94}
& 0.12 \\
\bottomrule
\end{tabular}%
}
\end{table*}

Training without target-empty queries produces the weakest results, including a mean Dice of \(75.46\), a mean HD95 of \(41.97\)~mm, and an FP-neg-rate of \(0.35\). Adding hard-negative or fully negative queries is associated with improved overlap, surface agreement, and false-positive control. The fixed primary configuration, \(20/10/25/15\), yields a mean Dice of \(83.13\), a mean HD95 of \(8.20\)~mm, and the lowest FP-neg-rate in the table (\(0.10\)). Relative to this primary configuration, the \(30/15/35/20\) setting increases mean Dice by \(1.06\) percentage points and reduces mean HD95 by \(0.59\)~mm, although its FP-neg-rate is slightly higher. TC and, particularly, ET occupy only a limited subset of axial slices in a BraTS volume; many slices are therefore empty for the selected target region. The higher-ratio result is consistent with the possibility that more frequent exposure to target-absent queries better reflects this slice-wise composition. Accordingly, different balances of hard-negative and fully negative episodes are associated with substantial differences in aggregate volume-level Dice and HD95.

\subsection{Computational Complexity of Component Ablations}
\label{sec:ablation_complexity}

All complexity values in this subsection and the following subsection were profiled in inference mode with batch size one. Each image input is a four-channel \(4\times256\times256\) MRI slice. For support-conditioned methods, one measurement corresponds to a one-shot episode comprising one support image, its binary support mask, and one query image; methods without a labeled support input are profiled on one query image. The reported floating-point operations (FLOPs) and multiply--accumulate operations (MACs), reported in units of \(10^{9}\) as GFLOPs and GMACs, include only the neural-network forward pass and exclude flip-based test-time augmentation, probability-volume reconstruction, smoothing, thresholding, nested-region post-processing, and data input/output. Accordingly, they describe standardized model-side operation counts rather than end-to-end wall-clock latency.

\autoref{tab:ablation_complexity} reports the computational footprint of the full model and the principal component ablations under this common profiling convention. Removing DMP produces the largest reduction in parameters and operations, whereas PC also contributes materially to the total footprint. In contrast, removing SGB or BR changes the measured complexity only modestly.

\begin{table}[t!]
\caption{Model-side complexity of FSS-UBrain ablations per \(4\times256\times256\) one-shot support--query episode at batch size one. Test-time augmentation and volume post-processing are excluded. Lowest values are bold.}
\label{tab:ablation_complexity}
\centering
\footnotesize
\renewcommand{\arraystretch}{1.15}
\setlength{\tabcolsep}{3.5pt}
% \resizebox{0.8\columnwidth}{!}{%
\begin{tabular}{@{}l c c c@{}}
\toprule
\textbf{Model} & \textbf{Params (M)} & \textbf{FLOPs (G)} & \textbf{MACs (G)} \\
\midrule
Without DMP
& \textbf{19.83}
& \textbf{200.45}
& \textbf{100.22} \\
Without SGB    & 41.82 & 351.05 & 175.53 \\
Without PC     & 36.88 & 328.39 & 164.19 \\
Without BR     & 42.61 & 347.00 & 173.50 \\
FSS-UBrain (Full model) & 42.65 & 352.05 & 176.03 \\
\bottomrule
\end{tabular}%
% }
\end{table}

\subsection{Model-Level Complexity and Accuracy--Efficiency Trade-Off}
\label{sec:model_complexity_analysis}

\autoref{tab:model_complexity} places FSS-UBrain in the context of the compared architectures under the profiling convention stated above. For support-conditioned methods, the reported cost includes one support--query episode; for methods without a labeled support input, it corresponds to one query-slice forward pass. The values therefore make the different inference pathways explicit and should not be interpreted as direct end-to-end latency measurements.

FSS-UBrain has \(42.65\) million parameters, \(352.05\)~GFLOPs, and \(176.03\)~GMACs. Its parameter count is lower than those of ALPNet~\cite{ouyang2020self}, CATNet~\cite{lin2023few}, \mbox{Q-Net~\cite{shen2023q}}, VMUNet~\cite{ruan2024vm}, SAM~\cite{kirillov2023segment}, TransUNet~\cite{chen2021transunet}, and GMRD~\cite{cheng2024few}, while its operation count is lower than those of SAM~\cite{kirillov2023segment}, Q-Net~\cite{shen2023q}, and GMRD~\cite{cheng2024few}. Conversely, it requires more forward-pass computation than lightweight few-shot baselines such as PANet~\cite{wang2019panet} and ALPNet~\cite{ouyang2020self}. Because support-conditioned measurements include both a support pathway and a query pathway, whereas non-FSS measurements contain only a query pathway, these values should be interpreted as standardized model-side costs rather than directly comparable end-to-end latency. They nonetheless make the cost of support-conditioned adaptation explicit.

\begin{table}[t!]
\caption{Model-side complexity at batch size one. Support-conditioned methods are profiled per support--query episode and non-FSS methods per query slice; test-time augmentation and volume post-processing are excluded. Lowest values are bold separately for the non-FSS group (upper block, including SAM) and the support-conditioned group (lower block, including FSS-UBrain).}
\label{tab:model_complexity}
\centering
\footnotesize
\renewcommand{\arraystretch}{1.15}
\setlength{\tabcolsep}{3pt}
% \resizebox{\linewidth}{!}{
\begin{tabular}{@{}l c c c@{}}
\toprule
\textbf{Model} & \textbf{Params (M)} & \textbf{FLOPs (G)} & \textbf{MACs (G)} \\
\midrule
SwinUNet~\cite{cao2022swin}
& \textbf{6.94}
& \textbf{6.90}
& \textbf{3.45} \\
U-Net~\cite{ronneberger2015unet}      & 7.70   & 73.91  & 36.96 \\
UNet++~\cite{zhou2019unet++}          & 36.62  & 276.83 & 138.42 \\
TransUNet~\cite{chen2021transunet}    & 105.32 & 77.09  & 38.55 \\
VMUNet~\cite{ruan2024vm}              & 44.28  & 19.66  & 9.82 \\
\midrule
SAM~\cite{kirillov2023segment}        & 93.74  & 976.50 & 488.25 \\
\midrule
ALPNet~\cite{ouyang2020self}          & 43.03  & 180.09 & 90.50 \\
PANet~\cite{wang2019panet}
& \textbf{14.72}
& \textbf{102.07}
& \textbf{51.00} \\
CATNet~\cite{lin2023few}              & 45.80  & 191.33 & 95.66 \\
Q-Net~\cite{shen2023q}                & 46.13  & 533.00 & 266.50 \\
GMRD~\cite{cheng2024few}              & 100.03 & 574.01 & 287.01 \\
\midrule
FSS-UBrain (ours)                           & 42.65  & 352.05 & 176.03 \\
\bottomrule
\end{tabular}
% }
\end{table}

\section{Discussion}
\label{sec:discussion}

The results demonstrate that a single labeled support slice can provide effective region-specific guidance for volumetric WT, TC, and ET segmentation. Under the controlled volume-level evaluation protocol, FSS-UBrain achieves the highest Dice and lowest HD95 point estimates among the evaluated few-shot methods for all three regions on the held-out BraTS20 split. The consistently strong WT results indicate that the model can effectively exploit support information for a spatially extensive tumor region. The greater support-draw variation observed for TC and ET further highlights the importance of representative support selection for smaller and more heterogeneous subregions. Overall, these findings support region-conditioned one-shot inference for multi-region brain tumor segmentation.

The component-ablation results support the complementary roles of the proposed architectural modules. Removing descriptor matching produces the largest observed reduction in mean Dice, emphasizing the importance of representing heterogeneous foreground and background appearance through multiple support-derived descriptors. Removing the support-guided bottleneck produces the largest increase in mean HD95, showing that deep support--query interaction contributes substantially to boundary localization. The performance reductions observed after removing the pyramid-context or boundary-refinement module further indicate that contextual aggregation and boundary-aware correction complement descriptor matching and decoder-side reconstruction. Together, these results support the integrated FSS-UBrain architecture rather than reliance on any individual component.

The non-FSS comparison provides a broader contextual assessment across different supervision and inference paradigms. FSS-UBrain achieves the strongest WT and ET overlap and the lowest ET HD95 in this comparison, whereas SAM~\cite{kirillov2023segment} achieves the strongest TC Dice and lower WT and TC surface distances. These results show that FSS-UBrain is particularly effective for support-conditioned WT and ET delineation, while also illustrating the complementary strengths of fully supervised and support-conditioned approaches. Importantly, FSS-UBrain obtains these results by conditioning each prediction on one labeled support example, thereby providing explicit region-specific evidence without requiring target-domain parameter updates.

The cross-cohort experiments demonstrate that FSS-UBrain retains strong performance beyond the BraTS20 source cohort. Under target-supported evaluation, FSS-UBrain achieves the highest mean Dice and lowest finite-pair mean HD95 point estimates on both BraTS23 and BraTS-Africa among the compared few-shot methods. These results show that labeled target support can adapt the prediction process to cohort-specific image and tumor characteristics while keeping the learned model parameters fixed. The strong performance on BraTS-Africa is particularly relevant because this cohort introduces acquisition and population differences relative to the BraTS20 source cohort.

The matched-cohort BraTS-Africa experiment further emphasizes the importance of support provenance. For FSS-UBrain, replacing source support with target support increases the mean Dice point estimate from \(74.55\) to \(78.04\) and reduces the mean HD95 point estimate from \(13.03\) to \(11.03\)~mm under the same query cohort, model parameters, inference procedure, and post-processing protocol. Several comparison methods also exhibit favorable changes in at least one metric when target support is used, suggesting that support examples drawn from the target cohort can provide more cohort-aligned appearance and anatomical evidence for multiple architectures. However, the magnitude and direction of the changes remain method dependent, as also observed for CATNet~\cite{lin2023few}. The ability to benefit from target support therefore depends on how effectively each architecture represents and transfers support information.

These findings position support selection as a central element of the one-shot segmentation protocol. An informative support example should contain the selected region, provide a reliable annotation, and represent the appearance and boundary characteristics of the target cohort. This consideration is particularly important for TC and ET because their smaller extent and greater heterogeneity make their support-derived representations more sensitive to the selected case and slice. Carefully selected target-cohort support can therefore align one-shot inference with cohort-specific tumor appearance without modifying the trained parameters.

The negative-query experiments demonstrate the importance of explicitly representing target absence during episodic learning. Training without target-empty queries produces the weakest mean Dice and HD95 results and the highest FP-neg-rate. Hard-negative queries help the model distinguish the selected TC or ET region from other non-target tumor tissue, whereas fully negative queries provide direct supervision for suppressing foreground activation when no tumor tissue is present. Combining these query types therefore provides complementary evidence for foreground localization and background rejection.

The primary \(20/10/25/15\) configuration achieves the lowest FP-neg-rate, whereas the higher-ratio \(30/15/35/20\) configuration achieves the highest mean Dice and lowest mean HD95 with a slightly higher FP-neg-rate. This pattern reveals a meaningful balance between foreground recovery and conservative false-positive suppression. This trade-off also illustrates the complementary role of the reported metrics discussed in Section~\ref{sec:metrics_fairness}. Specificity remains within \(99.89\)--\(99.94\) across all six configurations in \autoref{tab:pos_only_neg_ratio}, including the configuration trained without any target-empty query, because BG voxels dominate the volume; it therefore offers little resolution among the configurations. The same behavior is expressed far more legibly by FP-neg-rate, which spans \(0.35\) to \(0.10\) over these configurations, a \(3.5\)-fold range. Reporting overlap, surface-distance, and target-absent metrics jointly is thus necessary to characterize the effect of negative-query sampling. More broadly, the sensitivity results show that the composition of positive, hard-negative, and fully negative episodes is an important design factor for region-wise FSS, especially for sparse tumor subregions that are absent from many axial slices.

Model-side profiling places these accuracy gains in the context of computational cost. FSS-UBrain requires \(42.65\) million parameters, \(352.05\)~GFLOPs, and \(176.03\)~GMACs per one-shot support--query episode. Its parameter count is lower than those of ALPNet~\cite{ouyang2020self}, CATNet~\cite{lin2023few}, Q-Net~\cite{shen2023q}, VMUNet~\cite{ruan2024vm}, SAM~\cite{kirillov2023segment}, TransUNet~\cite{chen2021transunet}, and GMRD~\cite{cheng2024few}. Its operation count is also lower than those of several computationally intensive compared architectures. The component-level profiling shows that descriptor matching accounts for the largest share of the additional computation, which is consistent with its substantial contribution to segmentation performance. Efficient descriptor construction and reuse of support representations therefore provide promising directions for reducing the computational cost of support-conditioned inference.

Several limitations also define opportunities for further development. First, FSS-UBrain operates on two-dimensional slices, whereas explicit through-plane or fully volumetric modeling could provide richer three-dimensional anatomical context. Second, one-shot performance depends on the representativeness and annotation quality of the selected support example. Multi-shot support aggregation, uncertainty-aware support weighting, and automatic support selection could improve robustness to support variation. Third, target-supported inference requires a small number of labeled target cases, which should be considered when deploying the method in new clinical cohorts. Fourth, descriptor-based adaptation and support-guided interaction introduce additional computational cost, motivating more efficient support encoding, descriptor construction, and representation reuse.

Finally, the present quantitative evaluation focuses on cohorts whose labels can be harmonized with the WT, TC, and ET definitions used in BraTS20. BraTS24 was excluded because its post-treatment ontology includes a distinct resection-cavity label \cite{de20242024}. Extending the framework to post-treatment imaging will therefore require task-specific modeling of the additional clinical labels rather than direct reuse of the pre-treatment composite regions.

\section{Conclusion}
\label{sec:conclusion}

This work introduced FSS-UBrain, a region-wise one-shot segmentation framework for WT, TC, and ET delineation from multimodal brain MRI. The method combines a shared residual encoder, support-derived FG and BG descriptors, support-guided bottleneck interaction, descriptor-gated decoding, and boundary-guided refinement. Episodic training additionally uses hard-negative and fully negative queries with empty-query regularization to reduce spurious foreground activation when the selected region is absent.

All principal comparisons were conducted after three-dimensional reconstruction, with validation-only threshold calibration and a fixed inference and post-processing protocol. On held-out BraTS20, FSS-UBrain achieves the highest Dice and lowest HD95 point estimates among the evaluated few-shot methods for WT, TC, and ET. Under target-supported cross-cohort evaluation, it also achieves the highest mean Dice and lowest finite-pair mean HD95 point estimates on BraTS23 and BraTS-Africa. Component ablations are consistent with complementary contributions from descriptor matching, support-guided interaction, contextual aggregation, and boundary refinement. Negative-query sensitivity analysis identifies a trade-off between overlap accuracy and false-positive suppression.

These findings are specific to a two-dimensional, target-supported one-shot setting and do not constitute evidence of annotation-free or fully volumetric generalization. BraTS24 was not included in the quantitative WT/TC/ET comparison because its post-treatment ontology contains a distinct resection-cavity label \cite{de20242024}.

This work suggests several research directions for future work. 
% Future work will investigate higher-dimensional support--query modeling. 
A first direction is a 2.5D extension that incorporates neighboring slices to provide through-plane context while retaining slice-level prediction. A further direction is a fully three-dimensional framework that constructs support descriptors from annotated subvolumes and conditions volumetric query features. These extensions would allow us to examine whether explicit inter-slice information improves the continuity of tumor predictions and the delineation of small or irregular regions. Such extensions will require consideration of the increased memory and computational costs, the spatial extent of support annotations, and the sampling of target-present and target-empty subvolumes. Multi-shot and uncertainty-aware support aggregation, together with efficient reuse of support representations, will also be investigated. Evaluation across independent training runs and broader clinical cohorts will be pursued, with task-specific label definitions for post-treatment imaging.
% \end{appendices}

\section*{CRediT authorship contribution statement}
{Truong Viet Vu:} Conceptualization, Methodology, Software, Validation, Investigation, Writing -- original draft. {Nguyen Phuc Nguyen:} Software, Validation, Data curation, Visualization. {Dang Thi Thu Hang:} Data curation, Investigation, Visualization. {Tran Thien Thanh:} Formal analysis, Writing -- review \& editing. {Vo Nguyen Quoc Bao:} Formal analysis, Writing -- review \& editing, Supervision. {Nguyen Thai Anh:} Conceptualization, Methodology, Writing -- review \& editing, Supervision, Project administration. {Ngo Hoang Tu:} Conceptualization, Methodology, Writing -- review \& editing, Supervision, Funding acquisition.

\section*{Declaration of competing interest}
The authors declare that they have no known competing financial interests or personal relationships that could have appeared to influence the work reported in this paper.

\section*{Acknowledgements}
This research is funded by Van Lang University, Vietnam under grant number VLU-2604-DT-VLT-KCT-GV-0108.

%References
\balance
\bibliographystyle{elsarticle-harv}
\bibliography{reference}

\begin{thebibliography}{52}
\expandafter\ifx\csname natexlab\endcsname\relax\def\natexlab#1{#1}\fi
\providecommand{\url}[1]{\texttt{#1}}
\providecommand{\href}[2]{#2}
\providecommand{\path}[1]{#1}
\providecommand{\DOIprefix}{doi:}
\providecommand{\ArXivprefix}{arXiv:}
\providecommand{\URLprefix}{URL: }
\providecommand{\Pubmedprefix}{pmid:}
\providecommand{\doi}[1]{\href{http://dx.doi.org/#1}{\path{#1}}}
\providecommand{\Pubmed}[1]{\href{pmid:#1}{\path{#1}}}
\providecommand{\bibinfo}[2]{#2}
\ifx\xfnm\relax \def\xfnm[#1]{\unskip,\space#1}\fi
%Type = Article
\bibitem[{Adewole et~al.(2025)Adewole, Rudie, Gbadamosi, Zhang, Raymond,
  Ajigbotoshso, Toyobo, Aguh, Omidiji, Akinola et~al.}]{adewole2025brats}
\bibinfo{author}{Adewole, M.}, \bibinfo{author}{Rudie, J.D.},
  \bibinfo{author}{Gbadamosi, A.}, \bibinfo{author}{Zhang, D.},
  \bibinfo{author}{Raymond, C.}, \bibinfo{author}{Ajigbotoshso, J.},
  \bibinfo{author}{Toyobo, O.}, \bibinfo{author}{Aguh, K.},
  \bibinfo{author}{Omidiji, O.}, \bibinfo{author}{Akinola, R.}, et~al.,
  \bibinfo{year}{2025}.
\newblock \bibinfo{title}{The {BraTS}-africa dataset: Expanding the brain tumor
  segmentation data to capture african populations}.
\newblock \bibinfo{journal}{Radiology Artif. Intell.} \bibinfo{volume}{7},
  \bibinfo{pages}{e240528}.
%Type = Article
\bibitem[{Bakas et~al.(2017)Bakas, Akbari, Sotiras, Bilello, Rozycki, Kirby,
  Freymann, Farahani and Davatzikos}]{bakas2017advancing}
\bibinfo{author}{Bakas, S.}, \bibinfo{author}{Akbari, H.},
  \bibinfo{author}{Sotiras, A.}, \bibinfo{author}{Bilello, M.},
  \bibinfo{author}{Rozycki, M.}, \bibinfo{author}{Kirby, J.S.},
  \bibinfo{author}{Freymann, J.B.}, \bibinfo{author}{Farahani, K.},
  \bibinfo{author}{Davatzikos, C.}, \bibinfo{year}{2017}.
\newblock \bibinfo{title}{Advancing the cancer genome atlas glioma {MRI}
  collections with expert segmentation labels and radiomic features}.
\newblock \bibinfo{journal}{Sci. Data} \bibinfo{volume}{4},
  \bibinfo{pages}{1--13}.
%Type = Article
\bibitem[{Bakas et~al.(2018)Bakas, Reyes, Jakab, Bauer, Rempfler, Crimi,
  Shinohara, Berger, Ha, Rozycki et~al.}]{bakas2018identifying}
\bibinfo{author}{Bakas, S.}, \bibinfo{author}{Reyes, M.},
  \bibinfo{author}{Jakab, A.}, \bibinfo{author}{Bauer, S.},
  \bibinfo{author}{Rempfler, M.}, \bibinfo{author}{Crimi, A.},
  \bibinfo{author}{Shinohara, R.T.}, \bibinfo{author}{Berger, C.},
  \bibinfo{author}{Ha, S.M.}, \bibinfo{author}{Rozycki, M.}, et~al.,
  \bibinfo{year}{2018}.
\newblock \bibinfo{title}{Identifying the best machine learning algorithms for
  brain tumor segmentation, progression assessment, and overall survival
  prediction in the {BraTS} challenge}.
\newblock \bibinfo{journal}{arXiv preprint arXiv:1811.02629} .
%Type = Inproceedings
\bibitem[{Cao et~al.(2022)Cao, Wang, Chen, Jiang, Zhang, Tian and
  Wang}]{cao2022swin}
\bibinfo{author}{Cao, H.}, \bibinfo{author}{Wang, Y.}, \bibinfo{author}{Chen,
  J.}, \bibinfo{author}{Jiang, D.}, \bibinfo{author}{Zhang, X.},
  \bibinfo{author}{Tian, Q.}, \bibinfo{author}{Wang, M.}, \bibinfo{year}{2022}.
\newblock \bibinfo{title}{{Swin-UNet}: {UNet-like} pure transformer for medical
  image segmentation}, in: \bibinfo{booktitle}{European Conference on Computer
  Vision}, \bibinfo{organization}{Springer}. pp. \bibinfo{pages}{205--218}.
%Type = Misc
\bibitem[{{Center for Biomedical Image Computing and Analytics (CBICA),
  University of Pennsylvania}(Accessed: Mar. 12, 2026)}]{brats2020data}
\bibinfo{author}{{Center for Biomedical Image Computing and Analytics (CBICA),
  University of Pennsylvania}}, \bibinfo{year}{Accessed: Mar. 12, 2026}.
\newblock \bibinfo{title}{Multimodal brain tumor segmentation challenge 2020:
  Data}.
\newblock \URLprefix \url{https://www.med.upenn.edu/cbica/brats2020/data.html}.
%Type = Article
\bibitem[{Chandrasekar et~al.(2026)Chandrasekar, Zhu, Radhika, Kamali and
  Subhashri}]{chandrasekar2026passivity}
\bibinfo{author}{Chandrasekar, A.}, \bibinfo{author}{Zhu, Q.},
  \bibinfo{author}{Radhika, T.}, \bibinfo{author}{Kamali, M.},
  \bibinfo{author}{Subhashri, A.R.}, \bibinfo{year}{2026}.
\newblock \bibinfo{title}{Passivity-based synchronization of {Markovian} jump
  inertial neural networks via adaptive event-driven protocol and its
  application to image encryption}.
\newblock \bibinfo{journal}{Acta Mathematica Scientia} \bibinfo{volume}{46},
  \bibinfo{pages}{1555--1580}.
%Type = Article
\bibitem[{Chen et~al.(2021)Chen, Lu, Yu, Luo, Adeli, Wang, Lu, Yuille and
  Zhou}]{chen2021transunet}
\bibinfo{author}{Chen, J.}, \bibinfo{author}{Lu, Y.}, \bibinfo{author}{Yu, Q.},
  \bibinfo{author}{Luo, X.}, \bibinfo{author}{Adeli, E.},
  \bibinfo{author}{Wang, Y.}, \bibinfo{author}{Lu, L.},
  \bibinfo{author}{Yuille, A.L.}, \bibinfo{author}{Zhou, Y.},
  \bibinfo{year}{2021}.
\newblock \bibinfo{title}{{TransUNet}: Transformers make strong encoders for
  medical image segmentation}.
\newblock \bibinfo{journal}{arXiv preprint arXiv:2102.04306} .
%Type = Article
\bibitem[{Chen et~al.(2017)Chen, Papandreou, Schroff and
  Adam}]{chen2017rethinking}
\bibinfo{author}{Chen, L.C.}, \bibinfo{author}{Papandreou, G.},
  \bibinfo{author}{Schroff, F.}, \bibinfo{author}{Adam, H.},
  \bibinfo{year}{2017}.
\newblock \bibinfo{title}{Rethinking atrous convolution for semantic image
  segmentation}.
\newblock \bibinfo{journal}{arXiv preprint arXiv:1706.05587} .
%Type = Article
\bibitem[{Cheng et~al.(2024)Cheng, Wang, Xin, Zhou, Zhang and
  Shao}]{cheng2024few}
\bibinfo{author}{Cheng, Z.}, \bibinfo{author}{Wang, S.}, \bibinfo{author}{Xin,
  T.}, \bibinfo{author}{Zhou, T.}, \bibinfo{author}{Zhang, H.},
  \bibinfo{author}{Shao, L.}, \bibinfo{year}{2024}.
\newblock \bibinfo{title}{Few-shot medical image segmentation via generating
  multiple representative descriptors}.
\newblock \bibinfo{journal}{IEEE Transactions on Medical Imaging}
  \bibinfo{volume}{43}, \bibinfo{pages}{2202--2214}.
%Type = Inproceedings
\bibitem[{{\c{C}}i{\c{c}}ek et~al.(2016){\c{C}}i{\c{c}}ek, Abdulkadir,
  Lienkamp, Brox and Ronneberger}]{cciccek20163d}
\bibinfo{author}{{\c{C}}i{\c{c}}ek, {\"O}.}, \bibinfo{author}{Abdulkadir, A.},
  \bibinfo{author}{Lienkamp, S.S.}, \bibinfo{author}{Brox, T.},
  \bibinfo{author}{Ronneberger, O.}, \bibinfo{year}{2016}.
\newblock \bibinfo{title}{{3D U-Net}: learning dense volumetric segmentation
  from sparse annotation}, in: \bibinfo{booktitle}{International Conference on
  Medical Image Computing and Computer-Assisted Intervention},
  \bibinfo{organization}{Springer}. pp. \bibinfo{pages}{424--432}.
%Type = Article
\bibitem[{Debnath et~al.(2025)Debnath, Rahman, Azam, Zhang and
  Jonkman}]{debnath2025fss}
\bibinfo{author}{Debnath, R.K.}, \bibinfo{author}{Rahman, M.A.},
  \bibinfo{author}{Azam, S.}, \bibinfo{author}{Zhang, Y.},
  \bibinfo{author}{Jonkman, M.}, \bibinfo{year}{2025}.
\newblock \bibinfo{title}{{FSS-ULivR}: a clinically-inspired few-shot
  segmentation framework for liver imaging using unified representations and
  attention mechanisms}.
\newblock \bibinfo{journal}{J. Cancer Res. Clin. Oncol.} \bibinfo{volume}{151},
  \bibinfo{pages}{1--23}.
%Type = Article
\bibitem[{Despotovi{\'c} et~al.(2015)Despotovi{\'c}, Goossens and
  Philips}]{despotovic2015mri}
\bibinfo{author}{Despotovi{\'c}, I.}, \bibinfo{author}{Goossens, B.},
  \bibinfo{author}{Philips, W.}, \bibinfo{year}{2015}.
\newblock \bibinfo{title}{{MRI} segmentation of the human brain: Challenges,
  methods, and applications}.
\newblock \bibinfo{journal}{Comput. Math. Methods Med.} \bibinfo{volume}{2015},
  \bibinfo{pages}{450341}.
%Type = Inproceedings
\bibitem[{Fan et~al.(2026)Fan, Jia, Xiao, Yu, Yang, Yang, Xu, Huang and
  Wang}]{fan2026spenet}
\bibinfo{author}{Fan, C.}, \bibinfo{author}{Jia, X.}, \bibinfo{author}{Xiao,
  A.}, \bibinfo{author}{Yu, H.}, \bibinfo{author}{Yang, Z.},
  \bibinfo{author}{Yang, D.}, \bibinfo{author}{Xu, H.}, \bibinfo{author}{Huang,
  Y.}, \bibinfo{author}{Wang, L.}, \bibinfo{year}{2026}.
\newblock \bibinfo{title}{{SPENet}: Self-guided prototype enhancement network
  for few-shot medical image segmentation}, in: \bibinfo{booktitle}{Medical
  Image Computing and Computer Assisted Intervention -- MICCAI 2025},
  \bibinfo{publisher}{Springer}. pp. \bibinfo{pages}{584--593}.
%Type = Article
\bibitem[{Han et~al.(2022)Han, Wang, Chen, Chen, Guo, Liu, Tang, Xiao, Xu, Xu
  et~al.}]{han2022survey}
\bibinfo{author}{Han, K.}, \bibinfo{author}{Wang, Y.}, \bibinfo{author}{Chen,
  H.}, \bibinfo{author}{Chen, X.}, \bibinfo{author}{Guo, J.},
  \bibinfo{author}{Liu, Z.}, \bibinfo{author}{Tang, Y.}, \bibinfo{author}{Xiao,
  A.}, \bibinfo{author}{Xu, C.}, \bibinfo{author}{Xu, Y.}, et~al.,
  \bibinfo{year}{2022}.
\newblock \bibinfo{title}{A survey on vision transformer}.
\newblock \bibinfo{journal}{IEEE Trans. Pattern Anal. Mach. Intell.}
  \bibinfo{volume}{45}, \bibinfo{pages}{87--110}.
%Type = Article
\bibitem[{Hartmann et~al.(2023)Hartmann, Schmid, Meyer and
  Kramer}]{hartmann2023mism}
\bibinfo{author}{Hartmann, D.}, \bibinfo{author}{Schmid, V.},
  \bibinfo{author}{Meyer, P.}, \bibinfo{author}{Kramer, F.},
  \bibinfo{year}{2023}.
\newblock \bibinfo{title}{{MISM}: A medical image segmentation metric for
  evaluation of weak labeled data}.
\newblock \bibinfo{journal}{Diagnostics} \bibinfo{volume}{13},
  \bibinfo{pages}{2618}.
%Type = Inproceedings
\bibitem[{Hatamizadeh et~al.(2021)Hatamizadeh, Nath, Tang, Yang, Roth and
  Xu}]{hatamizadeh2021swin}
\bibinfo{author}{Hatamizadeh, A.}, \bibinfo{author}{Nath, V.},
  \bibinfo{author}{Tang, Y.}, \bibinfo{author}{Yang, D.},
  \bibinfo{author}{Roth, H.R.}, \bibinfo{author}{Xu, D.}, \bibinfo{year}{2021}.
\newblock \bibinfo{title}{Swin-{UNETR}: Swin transformers for semantic
  segmentation of brain tumors in {MRI} images}, in: \bibinfo{booktitle}{Int.
  MICCAI Brainlesion Workshop}, \bibinfo{organization}{Springer}. pp.
  \bibinfo{pages}{272--284}.
%Type = Inproceedings
\bibitem[{Hatamizadeh et~al.(2022)Hatamizadeh, Tang, Nath, Yang, Myronenko,
  Landman, Roth and Xu}]{hatamizadeh2022unetr}
\bibinfo{author}{Hatamizadeh, A.}, \bibinfo{author}{Tang, Y.},
  \bibinfo{author}{Nath, V.}, \bibinfo{author}{Yang, D.},
  \bibinfo{author}{Myronenko, A.}, \bibinfo{author}{Landman, B.},
  \bibinfo{author}{Roth, H.R.}, \bibinfo{author}{Xu, D.}, \bibinfo{year}{2022}.
\newblock \bibinfo{title}{{UNETR}: Transformers for {3D} medical image
  segmentation}, in: \bibinfo{booktitle}{Proc. IEEE Winter Conf. Appl. Comput.
  Vis.}, pp. \bibinfo{pages}{574--584}.
%Type = Inproceedings
\bibitem[{He et~al.(2016)He, Zhang, Ren and Sun}]{he2016deep}
\bibinfo{author}{He, K.}, \bibinfo{author}{Zhang, X.}, \bibinfo{author}{Ren,
  S.}, \bibinfo{author}{Sun, J.}, \bibinfo{year}{2016}.
\newblock \bibinfo{title}{Deep residual learning for image recognition}, in:
  \bibinfo{booktitle}{Proceedings of the IEEE Conference on Computer Vision and
  Pattern Recognition}, pp. \bibinfo{pages}{770--778}.
%Type = Article
\bibitem[{Huang et~al.(2023)Huang, Chan, Lok, Zhang, Lin, Lucero-Prisno, Xu,
  Zheng, Elcarte, Withers et~al.}]{huang2023disease}
\bibinfo{author}{Huang, J.}, \bibinfo{author}{Chan, S.C.},
  \bibinfo{author}{Lok, V.}, \bibinfo{author}{Zhang, L.}, \bibinfo{author}{Lin,
  X.}, \bibinfo{author}{Lucero-Prisno, D.E.}, \bibinfo{author}{Xu, W.},
  \bibinfo{author}{Zheng, Z.J.}, \bibinfo{author}{Elcarte, E.},
  \bibinfo{author}{Withers, M.}, et~al., \bibinfo{year}{2023}.
\newblock \bibinfo{title}{Disease burden, risk factors, and trends of primary
  central nervous system {(CNS)} cancer: a global study of registries data}.
\newblock \bibinfo{journal}{Neuro-Oncol.} \bibinfo{volume}{25},
  \bibinfo{pages}{995--1005}.
%Type = Article
\bibitem[{Isensee et~al.(2021)Isensee, Jaeger, Kohl, Petersen and
  Maier-Hein}]{isensee2021nnu}
\bibinfo{author}{Isensee, F.}, \bibinfo{author}{Jaeger, P.F.},
  \bibinfo{author}{Kohl, S.A.}, \bibinfo{author}{Petersen, J.},
  \bibinfo{author}{Maier-Hein, K.H.}, \bibinfo{year}{2021}.
\newblock \bibinfo{title}{{nnU-Net}: a self-configuring method for deep
  learning-based biomedical image segmentation}.
\newblock \bibinfo{journal}{Nat. Methods} \bibinfo{volume}{18},
  \bibinfo{pages}{203--211}.
%Type = Article
\bibitem[{Jiang et~al.(2022)Jiang, Zhang, Lin, Dong, Cheng and
  Liang}]{jiang2022swinbts}
\bibinfo{author}{Jiang, Y.}, \bibinfo{author}{Zhang, Y.}, \bibinfo{author}{Lin,
  X.}, \bibinfo{author}{Dong, J.}, \bibinfo{author}{Cheng, T.},
  \bibinfo{author}{Liang, J.}, \bibinfo{year}{2022}.
\newblock \bibinfo{title}{{SwinBTS}: A method for {3D} multimodal brain tumor
  segmentation using swin transformer}.
\newblock \bibinfo{journal}{Brain Sci.} \bibinfo{volume}{12},
  \bibinfo{pages}{797}.
%Type = Inproceedings
\bibitem[{Kirillov et~al.(2023)Kirillov, Mintun, Ravi, Mao, Rolland, Gustafson,
  Xiao, Whitehead, Berg, Lo et~al.}]{kirillov2023segment}
\bibinfo{author}{Kirillov, A.}, \bibinfo{author}{Mintun, E.},
  \bibinfo{author}{Ravi, N.}, \bibinfo{author}{Mao, H.},
  \bibinfo{author}{Rolland, C.}, \bibinfo{author}{Gustafson, L.},
  \bibinfo{author}{Xiao, T.}, \bibinfo{author}{Whitehead, S.},
  \bibinfo{author}{Berg, A.C.}, \bibinfo{author}{Lo, W.Y.}, et~al.,
  \bibinfo{year}{2023}.
\newblock \bibinfo{title}{Segment anything}, in:
  \bibinfo{booktitle}{Proceedings of the IEEE/CVF International Conference on
  Computer Vision}, pp. \bibinfo{pages}{4015--4026}.
%Type = Inproceedings
\bibitem[{Lee et~al.(2015)Lee, Xie, Gallagher, Zhang and Tu}]{lee2015deeply}
\bibinfo{author}{Lee, C.Y.}, \bibinfo{author}{Xie, S.},
  \bibinfo{author}{Gallagher, P.}, \bibinfo{author}{Zhang, Z.},
  \bibinfo{author}{Tu, Z.}, \bibinfo{year}{2015}.
\newblock \bibinfo{title}{Deeply-supervised nets}, in:
  \bibinfo{booktitle}{Artificial Intelligence and Statistics},
  \bibinfo{organization}{Pmlr}. pp. \bibinfo{pages}{562--570}.
%Type = Article
\bibitem[{Lin et~al.(2023a)Lin, Lin, Lu, Chen, Lin, Zhao, Shi, Qiu, Pan, Xu
  et~al.}]{lin2023ckd}
\bibinfo{author}{Lin, J.}, \bibinfo{author}{Lin, J.}, \bibinfo{author}{Lu, C.},
  \bibinfo{author}{Chen, H.}, \bibinfo{author}{Lin, H.}, \bibinfo{author}{Zhao,
  B.}, \bibinfo{author}{Shi, Z.}, \bibinfo{author}{Qiu, B.},
  \bibinfo{author}{Pan, X.}, \bibinfo{author}{Xu, Z.}, et~al.,
  \bibinfo{year}{2023}a.
\newblock \bibinfo{title}{{CKD-TransBTS}: clinical knowledge-driven hybrid
  transformer with modality-correlated cross-attention for brain tumor
  segmentation}.
\newblock \bibinfo{journal}{IEEE Trans. Med. Imaging} \bibinfo{volume}{42},
  \bibinfo{pages}{2451--2461}.
%Type = Inproceedings
\bibitem[{Lin et~al.(2017)Lin, Goyal, Girshick, He and
  Doll{\'a}r}]{lin2017focal}
\bibinfo{author}{Lin, T.Y.}, \bibinfo{author}{Goyal, P.},
  \bibinfo{author}{Girshick, R.}, \bibinfo{author}{He, K.},
  \bibinfo{author}{Doll{\'a}r, P.}, \bibinfo{year}{2017}.
\newblock \bibinfo{title}{Focal loss for dense object detection}, in:
  \bibinfo{booktitle}{Proceedings of the IEEE International Conference on
  Computer Vision}, pp. \bibinfo{pages}{2980--2988}.
%Type = Inproceedings
\bibitem[{Lin et~al.(2023b)Lin, Chen, Cheng and Chen}]{lin2023few}
\bibinfo{author}{Lin, Y.}, \bibinfo{author}{Chen, Y.}, \bibinfo{author}{Cheng,
  K.T.}, \bibinfo{author}{Chen, H.}, \bibinfo{year}{2023}b.
\newblock \bibinfo{title}{Few shot medical image segmentation with cross
  attention transformer}, in: \bibinfo{booktitle}{International Conference on
  Medical Image Computing and Computer-Assisted Intervention},
  \bibinfo{organization}{Springer}. pp. \bibinfo{pages}{233--243}.
%Type = Inproceedings
\bibitem[{Liu et~al.(2021)Liu, Lin, Cao, Hu, Wei, Zhang, Lin and
  Guo}]{liu2021swin}
\bibinfo{author}{Liu, Z.}, \bibinfo{author}{Lin, Y.}, \bibinfo{author}{Cao,
  Y.}, \bibinfo{author}{Hu, H.}, \bibinfo{author}{Wei, Y.},
  \bibinfo{author}{Zhang, Z.}, \bibinfo{author}{Lin, S.}, \bibinfo{author}{Guo,
  B.}, \bibinfo{year}{2021}.
\newblock \bibinfo{title}{Swin transformer: Hierarchical vision transformer
  using shifted windows}, in: \bibinfo{booktitle}{Proc. IEEE Int. Conf. Comput.
  Vis.}, pp. \bibinfo{pages}{10012--10022}.
%Type = Article
\bibitem[{Menze et~al.(2015)Menze, Jakab, Bauer, Kalpathy-Cramer, Farahani,
  Kirby, Burren, Porz, Slotboom, Wiest et~al.}]{menze2018multimodal}
\bibinfo{author}{Menze, B.}, \bibinfo{author}{Jakab, A.},
  \bibinfo{author}{Bauer, S.}, \bibinfo{author}{Kalpathy-Cramer, J.},
  \bibinfo{author}{Farahani, K.}, \bibinfo{author}{Kirby, J.},
  \bibinfo{author}{Burren, Y.}, \bibinfo{author}{Porz, N.},
  \bibinfo{author}{Slotboom, J.}, \bibinfo{author}{Wiest, R.}, et~al.,
  \bibinfo{year}{2015}.
\newblock \bibinfo{title}{The multimodal brain tumor image segmentation
  benchmark ({BraTS})}.
\newblock \bibinfo{journal}{IEEE Trans. Med. Imaging} \bibinfo{volume}{34},
  \bibinfo{pages}{1993--2024}.
%Type = Inproceedings
\bibitem[{Milletari et~al.(2016)Milletari, Navab and Ahmadi}]{milletari2016v}
\bibinfo{author}{Milletari, F.}, \bibinfo{author}{Navab, N.},
  \bibinfo{author}{Ahmadi, S.A.}, \bibinfo{year}{2016}.
\newblock \bibinfo{title}{{V-Net}: Fully convolutional neural networks for
  volumetric medical image segmentation}, in: \bibinfo{booktitle}{2016 Fourth
  International Conference on 3D Vision (3DV)}, \bibinfo{organization}{Ieee}.
  pp. \bibinfo{pages}{565--571}.
%Type = Inproceedings
\bibitem[{Min et~al.(2021)Min, Kang and Cho}]{min2021hypercorrelation}
\bibinfo{author}{Min, J.}, \bibinfo{author}{Kang, D.}, \bibinfo{author}{Cho,
  M.}, \bibinfo{year}{2021}.
\newblock \bibinfo{title}{Hypercorrelation squeeze for few-shot segmentation},
  in: \bibinfo{booktitle}{Proceedings of the IEEE/CVF International Conference
  on Computer Vision}, pp. \bibinfo{pages}{6941--6952}.
%Type = Inproceedings
\bibitem[{Nguyen et~al.(2025)Nguyen, Hang, Vu, Tu and Anh}]{nguyen2025swin}
\bibinfo{author}{Nguyen, N.P.}, \bibinfo{author}{Hang, D.T.T.},
  \bibinfo{author}{Vu, T.V.}, \bibinfo{author}{Tu, N.H.}, \bibinfo{author}{Anh,
  N.T.}, \bibinfo{year}{2025}.
\newblock \bibinfo{title}{{Swin-PResU}: Hybrid swin {ViT} and pruned residual
  {CNN} for precise brain tumor {MRI} segmentation}, in:
  \bibinfo{booktitle}{2025 RIVF International Conference on Computing and
  Communication Technologies (RIVF)}, \bibinfo{organization}{IEEE}. pp.
  \bibinfo{pages}{149--154}.
%Type = Article
\bibitem[{Oktay et~al.(2018)Oktay, Schlemper, Folgoc, Lee, Heinrich, Misawa,
  Mori, McDonagh, Hammerla, Kainz et~al.}]{oktay2018attention}
\bibinfo{author}{Oktay, O.}, \bibinfo{author}{Schlemper, J.},
  \bibinfo{author}{Folgoc, L.L.}, \bibinfo{author}{Lee, M.},
  \bibinfo{author}{Heinrich, M.}, \bibinfo{author}{Misawa, K.},
  \bibinfo{author}{Mori, K.}, \bibinfo{author}{McDonagh, S.},
  \bibinfo{author}{Hammerla, N.Y.}, \bibinfo{author}{Kainz, B.}, et~al.,
  \bibinfo{year}{2018}.
\newblock \bibinfo{title}{Attention {U-Net}: Learning where to look for the
  pancreas}.
\newblock \bibinfo{journal}{arXiv preprint arXiv:1804.03999} .
%Type = Inproceedings
\bibitem[{Ouyang et~al.(2020)Ouyang, Biffi, Chen, Kart, Qiu and
  Rueckert}]{ouyang2020self}
\bibinfo{author}{Ouyang, C.}, \bibinfo{author}{Biffi, C.},
  \bibinfo{author}{Chen, C.}, \bibinfo{author}{Kart, T.}, \bibinfo{author}{Qiu,
  H.}, \bibinfo{author}{Rueckert, D.}, \bibinfo{year}{2020}.
\newblock \bibinfo{title}{Self-supervision with superpixels: Training few-shot
  medical image segmentation without annotation}, in:
  \bibinfo{booktitle}{European Conference on Computer Vision},
  \bibinfo{organization}{Springer}. pp. \bibinfo{pages}{762--780}.
%Type = Inproceedings
\bibitem[{Perez et~al.(2018)Perez, Strub, De~Vries, Dumoulin and
  Courville}]{perez2018film}
\bibinfo{author}{Perez, E.}, \bibinfo{author}{Strub, F.},
  \bibinfo{author}{De~Vries, H.}, \bibinfo{author}{Dumoulin, V.},
  \bibinfo{author}{Courville, A.}, \bibinfo{year}{2018}.
\newblock \bibinfo{title}{{FiLM}: Visual reasoning with a general conditioning
  layer}, in: \bibinfo{booktitle}{Proceedings of the AAAI Conference on
  Artificial Intelligence}, pp. \bibinfo{pages}{3942--3951}.
%Type = Article
\bibitem[{Radhika et~al.(2026)Radhika, Chandrasekar, Subhashri, Kamali and
  Ahmad}]{radhika2026memory}
\bibinfo{author}{Radhika, T.}, \bibinfo{author}{Chandrasekar, A.},
  \bibinfo{author}{Subhashri, A.R.}, \bibinfo{author}{Kamali, M.},
  \bibinfo{author}{Ahmad, H.}, \bibinfo{year}{2026}.
\newblock \bibinfo{title}{Memory-sampled data controller for exponential
  synchronization of {Markovian} jump neural networks with mixed delays and
  partially unknown transition probabilities}.
\newblock \bibinfo{journal}{Physica A: Statistical Mechanics and its
  Applications} \bibinfo{volume}{694}, \bibinfo{pages}{131587}.
%Type = Inproceedings
\bibitem[{Ronneberger et~al.(2015)Ronneberger, Fischer and
  Brox}]{ronneberger2015unet}
\bibinfo{author}{Ronneberger, O.}, \bibinfo{author}{Fischer, P.},
  \bibinfo{author}{Brox, T.}, \bibinfo{year}{2015}.
\newblock \bibinfo{title}{{U-Net}: Convolutional networks for biomedical image
  segmentation}, in: \bibinfo{booktitle}{Int. Conf. Med. Image Comput.
  Comput.-Assisted Intervention},
  \bibinfo{organization}{\hspace{-0.2cm}Springer}. pp.
  \bibinfo{pages}{234--241}.
%Type = Article
\bibitem[{Ruan et~al.(2025)Ruan, Li and Xiang}]{ruan2024vm}
\bibinfo{author}{Ruan, J.}, \bibinfo{author}{Li, J.}, \bibinfo{author}{Xiang,
  S.}, \bibinfo{year}{2025}.
\newblock \bibinfo{title}{{VM-UNet}: Vision {Mamba} {UNet} for medical image
  segmentation}.
\newblock \bibinfo{journal}{ACM Transactions on Multimedia Computing,
  Communications, and Applications} .
%Type = Article
\bibitem[{Shaban et~al.(2017)Shaban, Bansal, Liu, Essa and
  Boots}]{shaban2017one}
\bibinfo{author}{Shaban, A.}, \bibinfo{author}{Bansal, S.},
  \bibinfo{author}{Liu, Z.}, \bibinfo{author}{Essa, I.},
  \bibinfo{author}{Boots, B.}, \bibinfo{year}{2017}.
\newblock \bibinfo{title}{One-shot learning for semantic segmentation}.
\newblock \bibinfo{journal}{arXiv preprint arXiv:1709.03410} .
%Type = Inproceedings
\bibitem[{Shen et~al.(2023)Shen, Li, Jin and Liu}]{shen2023q}
\bibinfo{author}{Shen, Q.}, \bibinfo{author}{Li, Y.}, \bibinfo{author}{Jin,
  J.}, \bibinfo{author}{Liu, B.}, \bibinfo{year}{2023}.
\newblock \bibinfo{title}{{Q-Net}: Query-informed few-shot medical image
  segmentation}, in: \bibinfo{booktitle}{Proceedings of SAI Intelligent Systems
  Conference}, \bibinfo{organization}{Springer}. pp. \bibinfo{pages}{610--628}.
%Type = Misc
\bibitem[{Synapse(Accessed: Mar. 12, 2026)}]{brats2023portal}
\bibinfo{author}{Synapse}, \bibinfo{year}{Accessed: Mar. 12, 2026}.
\newblock \bibinfo{title}{{BraTS} 2023 challenge}.
\newblock \URLprefix \url{https://www.synapse.org/brats2023}.
%Type = Article
\bibitem[{Taha and Hanbury(2015)}]{taha2015metrics}
\bibinfo{author}{Taha, A.A.}, \bibinfo{author}{Hanbury, A.},
  \bibinfo{year}{2015}.
\newblock \bibinfo{title}{Metrics for evaluating {3D} medical image
  segmentation: analysis, selection, and tool}.
\newblock \bibinfo{journal}{BMC Med. Imaging} \bibinfo{volume}{15},
  \bibinfo{pages}{29}.
%Type = Article
\bibitem[{Tamil~Thendral et~al.(2026)Tamil~Thendral, Ganesh~Babu, Chandrasekar
  and Cao}]{tamilthendral2026synchronization}
\bibinfo{author}{Tamil~Thendral, M.}, \bibinfo{author}{Ganesh~Babu, T.R.},
  \bibinfo{author}{Chandrasekar, A.}, \bibinfo{author}{Cao, Y.},
  \bibinfo{year}{2026}.
\newblock \bibinfo{title}{Synchronization of {Markovian} jump neural networks
  for sampled data control systems with additive delay components: Analysis of
  image encryption technique}.
\newblock \bibinfo{journal}{Mathematical Methods in the Applied Sciences}
  \bibinfo{volume}{49}, \bibinfo{pages}{1879--1895}.
%Type = Article
\bibitem[{Tian et~al.(2020)Tian, Zhao, Shu, Yang, Li and Jia}]{tian2020prior}
\bibinfo{author}{Tian, Z.}, \bibinfo{author}{Zhao, H.}, \bibinfo{author}{Shu,
  M.}, \bibinfo{author}{Yang, Z.}, \bibinfo{author}{Li, R.},
  \bibinfo{author}{Jia, J.}, \bibinfo{year}{2020}.
\newblock \bibinfo{title}{Prior guided feature enrichment network for few-shot
  segmentation}.
\newblock \bibinfo{journal}{IEEE Transactions on Pattern Analysis and Machine
  intelligence} \bibinfo{volume}{44}, \bibinfo{pages}{1050--1065}.
%Type = Article
\bibitem[{Vaswani et~al.(2017)Vaswani, Shazeer, Parmar, Uszkoreit, Jones,
  Gomez, Kaiser and Polosukhin}]{vaswani2017attention}
\bibinfo{author}{Vaswani, A.}, \bibinfo{author}{Shazeer, N.},
  \bibinfo{author}{Parmar, N.}, \bibinfo{author}{Uszkoreit, J.},
  \bibinfo{author}{Jones, L.}, \bibinfo{author}{Gomez, A.N.},
  \bibinfo{author}{Kaiser, {\L}.}, \bibinfo{author}{Polosukhin, I.},
  \bibinfo{year}{2017}.
\newblock \bibinfo{title}{Attention is all you need}.
\newblock \bibinfo{journal}{Advances in Neural Information Processing Systems}
  \bibinfo{volume}{30}.
%Type = Article
\bibitem[{de~Verdier et~al.(2024)de~Verdier, Saluja, Gagnon, LaBella, Baid,
  Tahon, Foltyn-Dumitru, Zhang, Alafif, Baig et~al.}]{de20242024}
\bibinfo{author}{de~Verdier, M.C.}, \bibinfo{author}{Saluja, R.},
  \bibinfo{author}{Gagnon, L.}, \bibinfo{author}{LaBella, D.},
  \bibinfo{author}{Baid, U.}, \bibinfo{author}{Tahon, N.H.},
  \bibinfo{author}{Foltyn-Dumitru, M.}, \bibinfo{author}{Zhang, J.},
  \bibinfo{author}{Alafif, M.}, \bibinfo{author}{Baig, S.}, et~al.,
  \bibinfo{year}{2024}.
\newblock \bibinfo{title}{The 2024 brain tumor segmentation ({BraTS})
  challenge: Glioma segmentation on post-treatment {MRI}}.
\newblock \bibinfo{journal}{arXiv preprint arXiv:2405.18368} .
%Type = Inproceedings
\bibitem[{Wang et~al.(2019)Wang, Liew, Zou, Zhou and Feng}]{wang2019panet}
\bibinfo{author}{Wang, K.}, \bibinfo{author}{Liew, J.H.}, \bibinfo{author}{Zou,
  Y.}, \bibinfo{author}{Zhou, D.}, \bibinfo{author}{Feng, J.},
  \bibinfo{year}{2019}.
\newblock \bibinfo{title}{{PANet}: Few-shot image semantic segmentation with
  prototype alignment}, in: \bibinfo{booktitle}{Proc. IEEE/CVF Int. Conf.
  Comput. Vis. (ICCV)}, pp. \bibinfo{pages}{9197--9206}.
%Type = Inproceedings
\bibitem[{Wu and He(2018)}]{wu2018group}
\bibinfo{author}{Wu, Y.}, \bibinfo{author}{He, K.}, \bibinfo{year}{2018}.
\newblock \bibinfo{title}{Group normalization}, in:
  \bibinfo{booktitle}{Proceedings of the European Conference on Computer Vision
  (ECCV)}, pp. \bibinfo{pages}{3--19}.
%Type = Inproceedings
\bibitem[{Xie et~al.(2021)Xie, Zhang, Shen and Xia}]{xie2021cotr}
\bibinfo{author}{Xie, Y.}, \bibinfo{author}{Zhang, J.}, \bibinfo{author}{Shen,
  C.}, \bibinfo{author}{Xia, Y.}, \bibinfo{year}{2021}.
\newblock \bibinfo{title}{{Cotr}: Efficiently bridging cnn and transformer for
  3d medical image segmentation}, in: \bibinfo{booktitle}{Int. Conf. Med. Image
  Comput. Comput.-Assist. Interv.}, \bibinfo{organization}{Springer}. pp.
  \bibinfo{pages}{171--180}.
%Type = Article
\bibitem[{Xu et~al.(2026)Xu, Zhu, Wang, Long and Zhang}]{xu2026bepmi}
\bibinfo{author}{Xu, B.}, \bibinfo{author}{Zhu, Y.}, \bibinfo{author}{Wang,
  S.}, \bibinfo{author}{Long, Y.}, \bibinfo{author}{Zhang, H.},
  \bibinfo{year}{2026}.
\newblock \bibinfo{title}{Few-shot medical image segmentation via
  boundary-extended prototypes and momentum inference}.
\newblock \bibinfo{journal}{Computer Vision and Image Understanding}
  \bibinfo{volume}{263}, \bibinfo{pages}{104571}.
%Type = Article
\bibitem[{Zhao et~al.(2025)Zhao, He, Yang, Geng, Zhu and Tang}]{zhao2025global}
\bibinfo{author}{Zhao, X.}, \bibinfo{author}{He, M.}, \bibinfo{author}{Yang,
  R.}, \bibinfo{author}{Geng, N.}, \bibinfo{author}{Zhu, X.},
  \bibinfo{author}{Tang, N.}, \bibinfo{year}{2025}.
\newblock \bibinfo{title}{The global, regional, and national brain and central
  nervous system cancer burden and trends from 1990 to 2021: An analysis based
  on the global burden of disease study 2021}.
\newblock \bibinfo{journal}{Front. Neurol.} \bibinfo{volume}{16},
  \bibinfo{pages}{1574614}.
%Type = Article
\bibitem[{Zhou et~al.(2021)Zhou, Guo, Zhang, Yu, Wang and
  Yu}]{zhou2022nnFormer}
\bibinfo{author}{Zhou, H.Y.}, \bibinfo{author}{Guo, J.},
  \bibinfo{author}{Zhang, Y.}, \bibinfo{author}{Yu, L.}, \bibinfo{author}{Wang,
  L.}, \bibinfo{author}{Yu, Y.}, \bibinfo{year}{2021}.
\newblock \bibinfo{title}{{nnFormer}: Interleaved transformer for volumetric
  segmentation}.
\newblock \bibinfo{journal}{arXiv preprint arXiv:2109.03201} .
%Type = Article
\bibitem[{Zhou et~al.(2019)Zhou, Siddiquee, Tajbakhsh and
  Liang}]{zhou2019unet++}
\bibinfo{author}{Zhou, Z.}, \bibinfo{author}{Siddiquee, M.M.R.},
  \bibinfo{author}{Tajbakhsh, N.}, \bibinfo{author}{Liang, J.},
  \bibinfo{year}{2019}.
\newblock \bibinfo{title}{{UNet++}: Redesigning skip connections to exploit
  multiscale features in image segmentation}.
\newblock \bibinfo{journal}{IEEE Transactions on Medical Imaging}
  \bibinfo{volume}{39}, \bibinfo{pages}{1856--1867}.

\end{thebibliography}

%%%%%%%%%%%%%%%%%%%%%%%%%  AUTHOR BIOGRAPHIES  (begin)  %%%%%%%%%%%%%%%%%%%%

%% Photo column width used by \authorbio.
\newlength{\biophotowidth}
\setlength{\biophotowidth}{0.95in}

%% \authorbio{photo file}{Author Name}{biography text}
\newcommand{\authorbio}[3]{%
  \begin{wrapfigure}{l}{\biophotowidth}%
    \vspace{-\intextsep}%
    \includegraphics[width=\biophotowidth]{#1}%
    \vspace{-0.6\baselineskip}%
  \end{wrapfigure}%
  \noindent\textbf{#2} #3\par\vspace{\baselineskip}%
}

\section*{Author biographies}

\authorbio{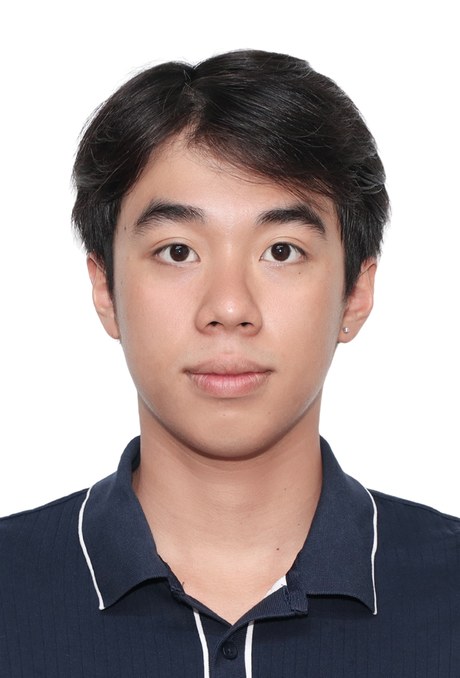}{Truong Viet Vu}{received the B.S. degree in information technology from the Van Lang School of Technology, Van Lang University, Ho Chi Minh City, Vietnam, in 2026. Since September 2025, he has been a member of the Future Communications and Computer Science (FCCS) Laboratory, Van Lang University. His research interests include machine learning, deep learning, and computer vision.}

\authorbio{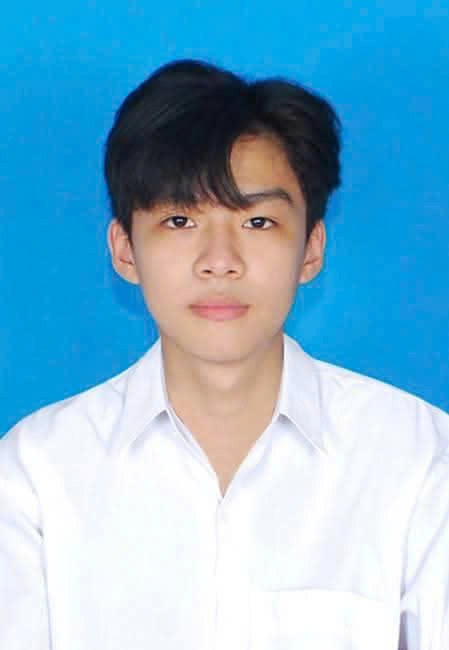}{Nguyen Phuc Nguyen}{received the B.S. degree in information technology from the Van Lang School of Technology, Van Lang University, Ho Chi Minh City, Vietnam, in 2026. Since September 2025, he has been a member of the Future Communications and Computer Science (FCCS) Laboratory, Van Lang University. His research interests include artificial intelligence, medical image segmentation, deep embedded clustering, and reinforcement learning.}

\authorbio{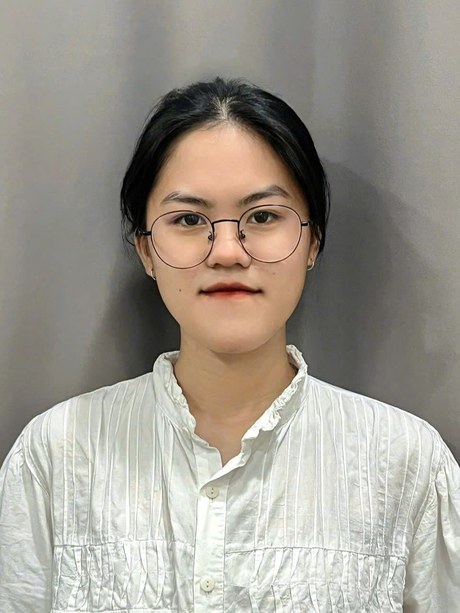}{Dang Thi Thu Hang}{received the B.S. degree in information technology from the Van Lang School of Technology, Van Lang University, Ho Chi Minh City, Vietnam, in 2026. Since September 2025, she has been a member of the Future Communications and Computer Science (FCCS) Laboratory, Van Lang University. Her research interests include machine learning, deep learning, big data, and data management systems.}

\authorbio{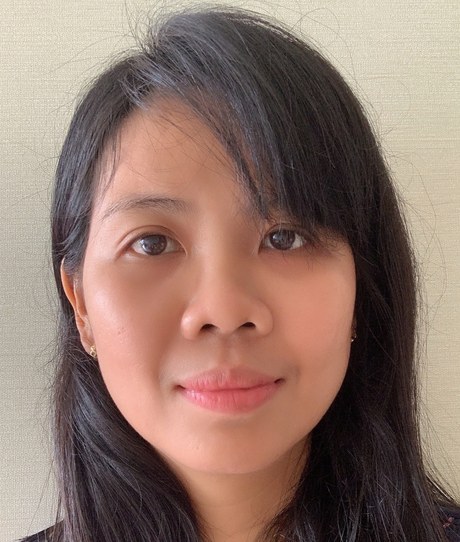}{Tran Thien Thanh}{received the Ph.D. degree in electrical engineering from Ho Chi Minh City University of Technology, Vietnam, in 2016. She serves as the Deputy Director of the Institute of Information Technology and Electrical-Electronics Engineering, Ho Chi Minh City University of Transport, Ho Chi Minh City, Vietnam. Her research interests include wireless communications and information theory with current emphasis on cooperative and cognitive communications, physical layer security, energy harvesting, 6G infrastructures, and applied machine learning.}

\authorbio{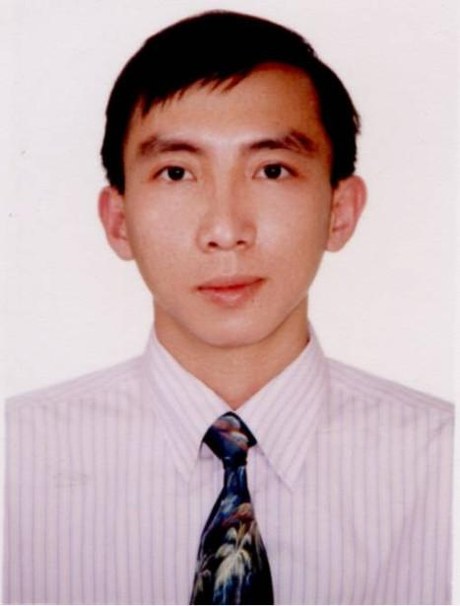}{Vo Nguyen Quoc Bao}{received the bachelor's and master's degrees from Ho Chi Minh City University of Technology, in 2002 and 2005, respectively, and the Ph.D. degree from the University of Ulsan, South Korea, in 2010. He is currently a Professor at Van Lang University, specializing in wireless communications. After completing his Ph.D., he was at the Posts and Telecommunications Institute of Technology, where he was promoted to an Associate Professor, in 2015, and a Professor, in 2023. In 2024, he joined Van Lang University, as a Full Professor. His research interests include developing new techniques at the physical layer for 5G and 6G communication systems and applied machine learning. He served as the Technical Editor-in-Chief for REV Journal on Electronics and Communications, from 2017 to 2023. He has also organized many IEEE conferences in Vietnam, including ATC, RIVF, ComManTel, and NICS.}

\authorbio{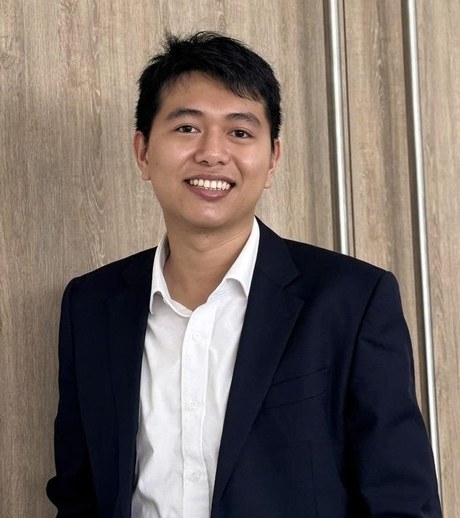}{Nguyen Thai Anh}{received the B.S. degree in information technology from the Van Lang School of Technology, Van Lang University (VLU), Ho Chi Minh City, Vietnam, in 2022, and the M.Sc. degree in data science and artificial intelligence from the School of Engineering and Technology, Asian Institute of Technology, Pathum Thani, Thailand, in 2024. From 2020 to 2021, he was a Data Scientist Intern with DMSpro. From 2021 to 2022, he was a Teaching Assistant with the Faculty of Information Technology, VLU, where he later served as an International Cooperation Specialist, from 2023 to 2024. Since 2024, he has been a Lecturer with the Faculty of Information Technology and a member of the Data Science Research Group, VLU. He is also a Guest Lecturer with the University of Industry and Trade, Ho Chi Minh City. His research interests include machine learning, deep learning, and computer vision.}

\authorbio{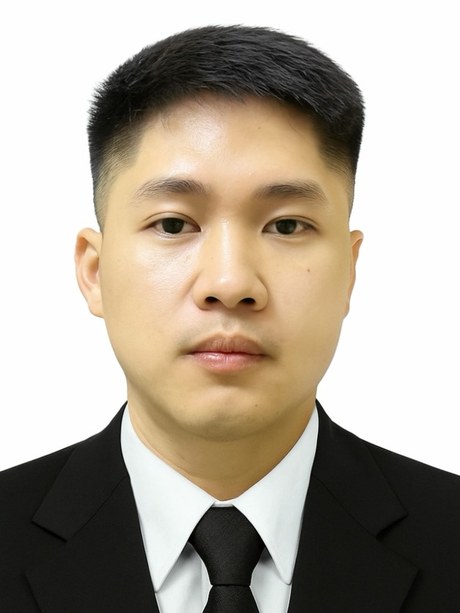}{Ngo Hoang Tu}{received the B.S. degree in computer networking and data communications from Ho Chi Minh City University of Transport (UTH), Ho Chi Minh City, Vietnam, in 2020, the M.S. degree from the Department of Smart Energy System Engineering, Seoul National University of Science and Technology (SeoulTech), Seoul, South Korea, in 2022, and the Ph.D. degree from the Department of Electrical and Information Engineering, SeoulTech, in 2025. From 2019 to 2020, he was an Assistant Researcher with the Wireless Communication Laboratory, Posts and Telecommunications Institute of Technology, Ho Chi Minh City. In 2020, he was a Lecturer with the Department of Computer Engineering, UTH. Since September 2025, he has been a Lecturer with the Faculty of Information Technology, Van Lang School of Technology, Van Lang University, Ho Chi Minh City. His research interests include wireless communications, signal processing, and applied machine learning. He received the Best Paper Award at the International Conference on Intelligence Systems and Robotics for Sustainable Development (ISRSD) in 2026, the Best Dissertation Award in 2025, the Best Presentation Awards at the Annual Conference of Vietnamese Young Scientists (ACVYS), in 2022 and 2024, and the Outstanding Student Paper Award at the International Conference on Control, Automation and Systems (ICCAS) in 2023. He serves on the Editorial Board for EAI Endorsed Transactions on Industrial Networks and Intelligent Systems.}

%%%%%%%%%%%%%%%%%%%%%%%%%  AUTHOR BIOGRAPHIES  (end)  %%%%%%%%%%%%%%%%%%%%%%

\end{document}